\documentclass[11pt]{article}

\usepackage[preprint]{iclr2026/acl}

\usepackage{times}
\usepackage{latexsym}

\usepackage[T1]{fontenc}

\usepackage[utf8]{inputenc}

\usepackage{microtype}

\usepackage{inconsolata}

\usepackage{amsmath,amsfonts,bm}

\def\eqref#1{equation~\ref{#1}}

\def\1{\bm{1}}

\DeclareMathAlphabet{\mathsfit}{\encodingdefault}{\sfdefault}{m}{sl}
\SetMathAlphabet{\mathsfit}{bold}{\encodingdefault}{\sfdefault}{bx}{n}

\usepackage{hyperref}
\usepackage{url}
\usepackage[most]{tcolorbox}
\usepackage{booktabs,xcolor}
\usepackage{colortbl}    
\usepackage{amssymb}
\usepackage{amsfonts}
\usepackage{makecell}
\usepackage{enumitem}
\usepackage{graphicx}
\usepackage{caption}
\usepackage{geometry}
\usepackage{float}
\usepackage{wrapfig}

\usepackage{listings}
\definecolor{lightgrey}{rgb}{0.827, 0.827, 0.827}
\newgeometry{top=1.2in, textheight=8.7in, textwidth=6.5in, headheight=12pt, headsep=25pt, footskip=30pt}

\definecolor{avg5}{HTML}{F3FAF9} 
\definecolor{avg6}{HTML}{EDF7F6} 
\definecolor{avg7}{HTML}{EAF7FB} 
\definecolor{avg8}{HTML}{DCEEF6} 
\definecolor{avgcol}{RGB}{240,240,240} 
\newcommand{\mixgray}[2]{\cellcolor{#1!90!gray}#2}

\title{DeepResearchEval:
An Automated Framework for Deep Research Task Construction and Agentic Evaluation}

\author{
\textbf{Yibo Wang}$^{1,2}$, 
\textbf{Lei Wang}$^1$\footnotemark[2], 
\textbf{Yue Deng}$^1$, 
\textbf{Keming Wu}$^1$, 
\textbf{Yao Xiao}$^1$, 
\textbf{Huanjin Yao}$^2$\\ 
\textbf{Liwei Kang}$^1$,
\textbf{Hai Ye}$^1$, 
\textbf{Yongcheng Jing}$^2$, 
\textbf{Lidong Bing}$^1$
\\
$^1$Infinity Lab, Shanda Group\\ 
$^2$Nanyang Technological University
}

\newif\ifshowcomments
\showcommentstrue      

\usepackage{CJKutf8}

\begin{document}

\twocolumn[{%
   \renewcommand\twocolumn[1][]{#1}%
   \maketitle
   \vspace{-40pt}
   \begin{center}
    \centering
    \centerline{\texttt{yibowang0021@gmail.com}, \texttt{\{lei.wang, lidong.bing\}@shanda.com}}
    \vspace{4pt}
    \centerline{
    \includegraphics[height=1em]{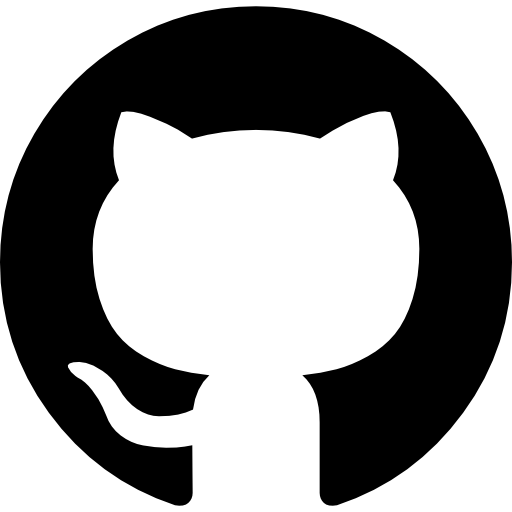}\ Code: \url{https://github.com/Infinity-AILab/DeepResearchEval}}
   \end{center}%
}]


\renewcommand{\thefootnote}{\fnsymbol{footnote}}
\footnotetext[2]{Corresponding Author}
\renewcommand{\thefootnote}{\arabic{footnote}}

\begin{abstract}

Deep research systems are widely used for multi-step web research, analysis, and cross-source synthesis, yet their evaluation remains challenging. 
Existing benchmarks often require annotation-intensive task construction, rely on static evaluation dimensions, or fail to reliably verify facts when citations are missing.
To bridge these gaps, we introduce \textbf{DeepResearchEval}, an automated framework for deep research task construction and agentic evaluation.
For task construction, we propose a persona-driven pipeline generating realistic, complex research tasks anchored in diverse user profiles, applying a two-stage filter (Task Qualification and Search Necessity) to retain only tasks requiring multi-source evidence integration and external retrieval.
For evaluation, we propose an agentic pipeline with two components: an \textit{Adaptive Point-wise Quality Evaluation} that dynamically derives task-specific evaluation dimensions, criteria, and weights conditioned on each generated task, and an \textit{Active Fact-Checking} that autonomously extracts and verifies report statements via web search, even when citations are missing. 

\end{abstract}

\begin{figure}[t!] 
  \centering
  \includegraphics[width=\linewidth]{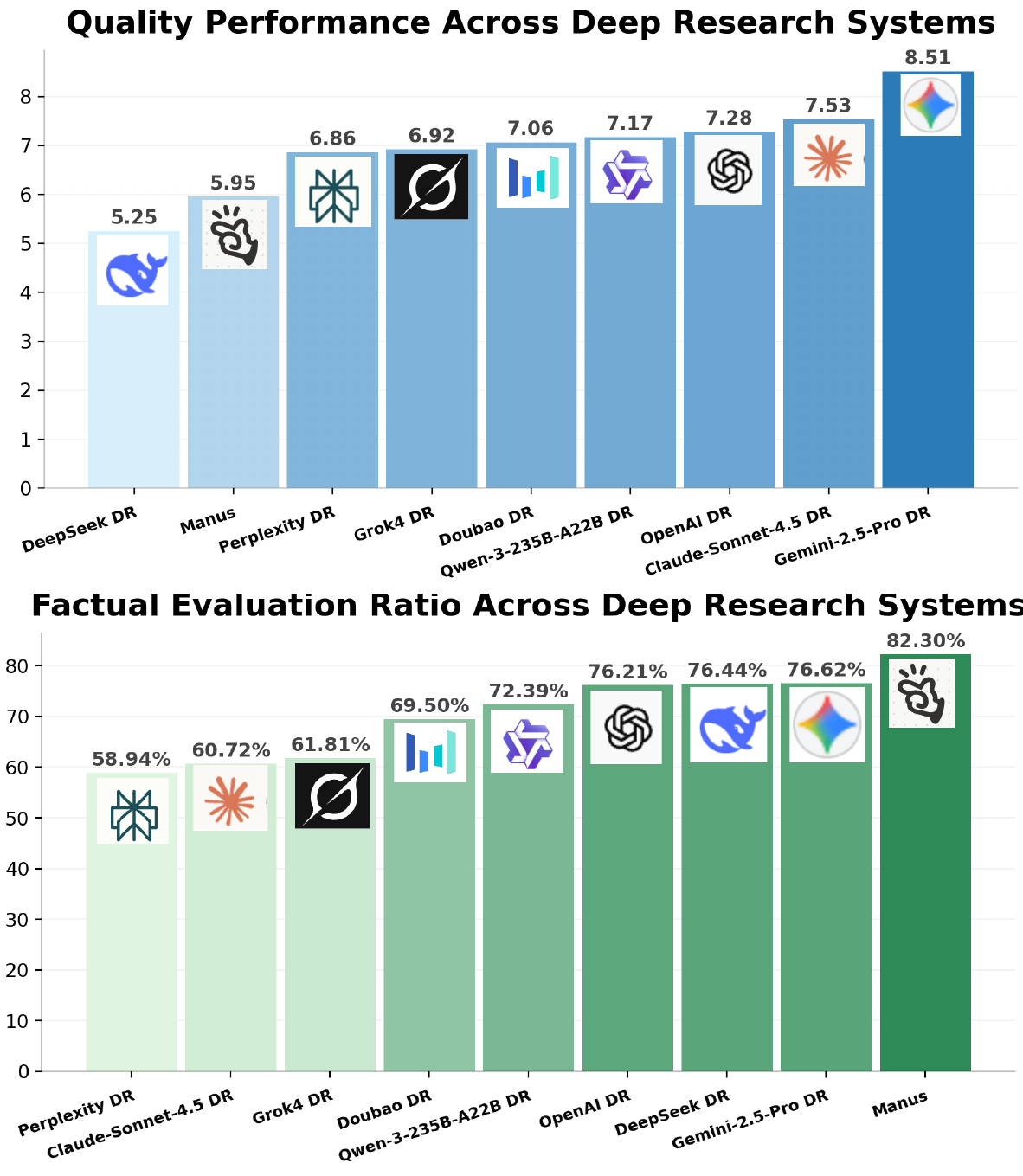}
  \caption{
Overview of deep research systems' performance on our benchmark.
The upper section reports quality evaluation results across deep research systems, with Gemini-2.5-Pro achieving the highest score ($8.51$/$10$).
The bottom section reports factual correctness, where Manus achieves the highest ratio of correct statements ($82.3\%$).
  }
  \label{fig:agent_performance_overview} 
  \vspace{-5mm}
\end{figure}

\section{Introduction}

The rapid advancement of Large Language Models (LLMs) has initiated a significant shift in AI capabilities, moving from passive text generation toward the development of agentic systems capable of tackling complex real-world tasks~\citep{liu2025deepseek, team2025kimi, zeng2025glm, deepmind_gemini3pro_modelcard_2025, openai2025gpt5systemcard}. Within this broader transition toward agentic intelligence, deep research systems have emerged as one of the representative paradigms~\citep{openai2025deepresearch, google2025deepresearch, tongyidr, perplexity2025deepresearch}. They autonomously conduct investigative processes that involve iterative web browsing, targeted information retrieval, cross-source verification, and multi-perspective synthesis. Through this structured workflow, deep research systems ultimately generate comprehensive, citation-grounded reports that traditionally require substantial human effort, e.g., ``Please review AI compute investments in 2025 and looking ahead to the trends of 2026''. 

With the transition of the technological paradigm, 
evaluating long reports generated by deep research systems poses a key challenge, as it differs substantially from conventional QA tasks.
Several benchmarks have been proposed to assess long-form, research-style outputs, however most existing benchmarks suffer from three limitations:
i) they rely on expert-driven task construction, which is annotation-intensive~\cite{yao2025rigorous,du2025deepresearch,abaskohi2025drbench,gou2025mind2web};
ii) they employ static quality evaluation dimensions shared across tasks~\cite{fan2025understanding,wang2025liveresearchbench}; 
and iii) they verify only citation-linked statements for factuality, leaving uncited factual claims unexamined~\cite{du2025deepresearch,fan2025understanding,gou2025mind2web}.

To bridge these gaps, we present \textbf{DeepResearchEval}, an automated framework for deep research task construction and agentic evaluation.
To address the scarcity of high-quality deep research tasks and the high cost of expert annotation, we propose a persona-driven pipeline, which constructs tasks anchored in specific personas, ensuring realistic needs and high complexity.
We then apply the Task Qualification Filter that assesses whether a task truly requires up-to-date evidence aggregation and multi-source investigation, and the Search Necessity Filter that discards simple questions solvable using an LLM's internal parametric knowledge. 
As a result, we retain 100 persona-driven, high-quality deep research tasks across multiple domains.

On the evaluation side, we develop an agentic evaluation pipeline with two key components.
(i) \emph{Adaptive Point-wise Quality Evaluation} preserves a fixed set of general evaluation dimensions shared across all tasks, while additionally generating task-specific dimensions, criteria, and corresponding weights for each task, allowing fine-grained and interpretable scoring tailored to individual deep research tasks.
(ii) \emph{Active Fact-Checking} performs {active} verification: it extracts verifiable statements (e.g., numbers, events, dates, entities), retrieves external evidence, and assigns structured labels (\texttt{Right}/\texttt{Wrong}/\texttt{Unknown}), thereby enabling factual verification of both cited and uncited claims.
While factual correctness is an inherently core component of overall report quality, we treat it as a standalone evaluation module due to its distinct verification process, which requires external evidence retrieval, and its critical role in ensuring the reliability of deep research reports.

We apply our framework to evaluate {leading deep research systems}, spanning proprietary general-purpose systems (e.g., Gemini Deep Research~\citep{google2025deepresearch}, OpenAI Deep Research~\citep{openai2025deepresearch}), and agentic generalists with deep research capabilities (e.g., Manus~\citep{manus2025}).
In total, we evaluate 900 deep research reports, comprising 100 tasks per system across 9 deep research systems.
As shown in Figure \ref{fig:agent_performance_overview}, our comprehensive evaluation reveals clear strengths and weaknesses across different dimensions of deep research capability. 
{Gemini Deep Research} achieve the strongest performance in {report quality evaluation}, reflecting superior coverage, insight, and structural coherence. {Manus} and {Gemini Deep Research} attain the highest scores in {factual evaluation}, showing stronger robustness against hallucinations and citation errors during complex multi-source report synthesis.
Additionally, we observe a consistent gap between general and task-specific evaluation: across all systems, task-specific scores are systematically lower than those on fixed general dimensions.
This gap indicates that current deep research systems often fail to meet task-specific success criteria, motivating task-adaptive evaluation beyond fixed general dimensions, precisely what our benchmark is designed to capture.
\section{Related Works}

\renewcommand{\arraystretch}{1.1}
\begin{table*}[!t]
    \centering
    \small
    
    \caption{Comparative Analysis of Deep Research Benchmarks.}
    \vspace{-2mm}
    \newcommand{\greencheck}{{\color{green!60!black}\checkmark}}
    \newcommand{\redx}{{\color{red!70!black}\texttimes}}

    \setlength{\tabcolsep}{4pt}
    \resizebox{0.99\linewidth}{!}{
    \begin{tabular}{l|ccccc}
        \toprule
        \textbf{Benchmark} &
        \textbf{\makecell{Automated Task\\Generation}} &
        \textbf{\makecell{Output\\Format}} & 
        \textbf{\makecell{Reference-free\\Evaluation}} & 
        \textbf{\makecell{Adaptive Evaluation\\Dimensions}} & 
        \textbf{\makecell{Active Fact\\Verification}} \\
        \midrule
        Mind2Web 2~\citep{gou2025mind2web} 
            & \redx & Report & \redx & \redx & \redx \\
        
        DeepResearch Bench~\citep{du2025deepresearch} 
            & \redx & Report & \redx & \redx & \redx \\
        
        ResearcherBench~\citep{xu2025researcherbench} 
            & \redx & Report & \greencheck & \redx & \redx \\
        
        BrowseComp-Plus~\citep{chen2025browsecomp} 
            & \redx & Short-Form Answer & \redx & \redx & \redx \\
        
        WideSearch~\citep{wong2025widesearch} 
            & \redx & Table-Style Answer & \redx & \redx & \redx \\
        
        ReportBench~\citep{li2025reportbench} 
            & \greencheck & Report & \redx & \redx & \redx \\
        
        DeepResearch Arena~\citep{wan2025deepresearch} 
            & \greencheck & Report & \greencheck & \redx & \redx \\
        
        DRBench~\citep{abaskohi2025drbench} 
            & \redx & Report & \redx & \redx & \redx \\
        
        LiveResearchBench~\citep{wang2025liveresearchbench} 
            & \redx & Report & \greencheck & \redx & \redx \\
        
        Finder~\citep{zhang2025far} 
            & \redx & Report & \greencheck & \redx & \redx \\
        
        \midrule
        \textbf{DeepResearchEval (Ours)} &
        \greencheck & Report & \greencheck &
        \greencheck & \greencheck \\
        \bottomrule
    \end{tabular}
    }
    \label{tab:deep_research_comparison}
    \vspace{-6mm}
\end{table*}

Deep research systems are a specialized class of agents designed for complex, multi-stage investigative tasks~\citep{openai2025deepresearch, google2025deepresearch, xai2025deepsearch, doubao2025deepresearch, perplexity2025deepresearch, manus2025, anthropic2025claude45}. 
Unlike conventional QA systems, they autonomously plan long-horizon workflows, navigate heterogeneous web sources, and synthesize information into structured, citation-grounded reports.
Existing systems broadly fall into proprietary solutions~\citep{openai2025deepresearch,google2025deepresearch,perplexity2025deepresearch} with limited transparency, and open-source efforts~\citep{li2025webthinker, zheng2025deepresearcher, qiao2025webresearcher} emphasizing modularity and reproducibility.

The emergence of deep research systems has motivated a broad range of benchmarks targeting different agentic capabilities~\citep{gou2025mind2web, mialon2023gaiabenchmarkgeneralai, phan2025humanity, du2025deepresearch, wong2025widesearch, wan2025deepresearch, abaskohi2025drbench, zhang2025far, wang2025liveresearchbench, li2025reportbench, chen2025browsecomp, wei2025browsecompsimplechallengingbenchmark, xu2025researcherbench,lei2025dacomp,luo2025ultrahorizonbenchmarkingagentcapabilities, yao2025rigorous,han2025deercomprehensivereliablebenchmark}. 
Early benchmarks such as GAIA~\citep{mialon2023gaiabenchmarkgeneralai} and Humanity’s Last Exam (HLE)~\citep{phan2025humanity} focus on general reasoning and tool use, while others emphasize persistent web navigation and retrieval, including WideSearch~\citep{wong2025widesearch} and BrowseComp variants~\citep{wei2025browsecompsimplechallengingbenchmark, chen2025browsecomp}. 
More recent benchmarks move toward report-level evaluation, including DeepResearch Bench~\citep{du2025deepresearch}, LiveResearchBench~\citep{wang2025liveresearchbench}, and DRBench~\citep{abaskohi2025drbench}; however, they remain annotation-intensive, rely on fixed task-agnostic evaluation dimensions, and often restrict factual verification to cited statements, leaving uncited claims unchecked.
In contrast, \textbf{DeepResearchEval} introduces an automated framework for task construction and agentic evaluation.
As summarized in Table~\ref{tab:deep_research_comparison}, it uniquely combines automatic task generation, reference-free evaluation, adaptive task-specific quality dimensions, and active fact verification over both cited and uncited statements for deep research systems.

\section{Task Construction}

Existing task collection relies heavily on expert annotators, suffering from three limitations:
(1) high-quality annotation  is costly and time-consuming;
(2) tasks are constrained by annotators’ individual backgrounds and domain knowledge; and
(3) task collection is static and difficult to update over time.

\begin{figure*}[t]
    \centering
    \includegraphics[width=0.96\linewidth]{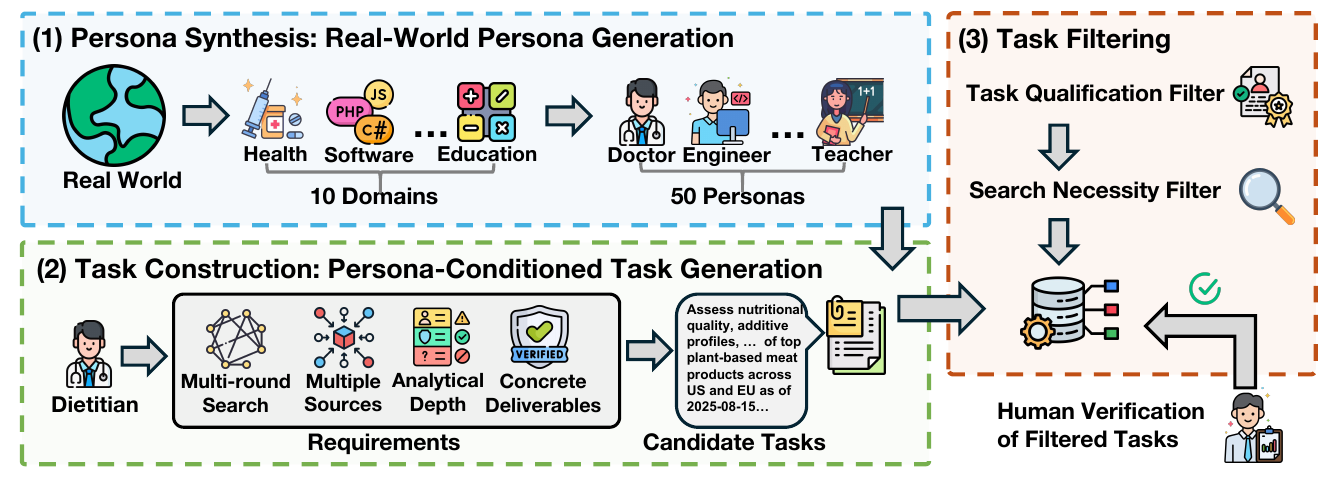}
    \vspace{-3mm}
    \caption{The proposed three-stage pipeline for constructing persona-driven deep research tasks.}
    \label{fig:task}
    \vspace{-3mm}
\end{figure*}

To address these limitations, we introduce an {automated} persona-driven deep research task collection pipeline that mirrors real-world production workflows.
As shown in Figure~\ref{fig:task}, we generate diverse personas conditioned on specific domains, which serve as seeds to produce expertise-aligned tasks, followed by multiple quality filtering stages to ensure a high-quality final set of deep research tasks.
Implementation details and prompts are provided in Appendix~\ref{apd:DR task}.

\subsection{Construction Pipeline}
\label{sec:pipeline}

\textbf{Persona Synthesis.}
To ensure our evaluation covers a diverse spectrum of real-world information needs, we draw upon the domain taxonomy defined in~\cite{wettig2025organize} and curate ten representative categories specifically suitable for deep research tasks, forming a domain set $\mathcal{D}$ that encompasses: \textit{Transportation}, \textit{Politics}, \textit{Finance \& Business}, \textit{History}, \textit{Software Development}, \textit{Industrial}, \textit{Sports \& Fitness}, \textit{Health}, \textit{Science \& Technology}, and \textit{Education \& Jobs}. 
For each domain $d \in \mathcal{D}$, we prompt an LLM to generate personas that are closely related to the domain while exhibiting diverse backgrounds. 
Each persona $p$ is specified by attributes including \textit{affiliation}, \textit{role}, \textit{background}, \textit{name}, and \textit{subdomain}. 
We generate five personas per domain, resulting in a total set $P$ of 50 personas. 

\textbf{Task Construction.}
For each persona $p \in P$, we prompt an LLM to generate candidate deep research tasks conditioned on the persona's background.
To ensure high task complexity, we enforce a generation schema requiring: (i) multi-round web searches; (ii) integration of evidence from diverse sources (e.g., papers, reports, and forums); (iii) sufficient analytical depth covering recent developments, data analysis, trend assessment, and comparative analysis; and (iv) concrete deliverables with explicit time constraints and 10--50 word descriptions. 
We generate four tasks per persona, yielding a total set of 200 candidates.

\textbf{Task Filtering.} 
To further ensure the quality of our benchmark, we employ a two-stage filtering pipeline for the candidate deep research tasks: a \textit{Task Qualification Filter} and a \textit{Search Necessity Filter}.

\textit{Task Qualification Filter}:
This distinguishes deep research tasks from simple tasks. An LLM-based evaluator assesses candidates on four criteria: requirement for up-to-date knowledge, multi-source evidence integration, multi-layered in-depth investigation, and persona's background and expertise alignment. Only tasks with a confidence score $> 0.7$ are retained.

\textit{Search Necessity Filter}:  
To exclude tasks solvable by internal knowledge, an LLM attempts each retained task $t$ using only parametric knowledge (no external tools). A separate evaluator assesses this non-search baseline across dimensions like accuracy, depth, and timeliness, professionalism, and structure. Tasks achieving high quality scores without search are filtered out, resulting in 155 retained tasks.

\begin{table}[t]
\centering
\caption{
Distribution of expert approval counts for the 155 retained tasks.}
\vspace{-2mm}
\label{tab:expert_agreement}
\resizebox{\linewidth}{!}{%
\begin{tabular}{c cccccccc}
\toprule
\textbf{Approval Count} 
& 0--1 & 2 & 3 & 4 & 5 & 6 & 7 \\
\midrule
\textbf{Proportion} 
& 0\% & 4\% & 15\% & 15\% & 26\% & 30\% & 9\% \\
\bottomrule
\end{tabular}
}
\vspace{-2mm}
\end{table}

\textbf{Human Verification of Retained Tasks.}
To validate the automated pipeline, we invited seven domain experts holding Ph.D. degrees to independently evaluate the $155$ retained tasks against deep research criteria, including multi-round search, multi-source evidence integration, and substantial analytical depth.
Table~\ref{tab:expert_agreement} shows that $80\%$ of tasks were deemed qualified  by at least four experts.
These results indicate the automated pipeline's effectiveness in reliably producing high-quality tasks.

\begin{figure}[t]
    \centering
    \includegraphics[width=\linewidth]{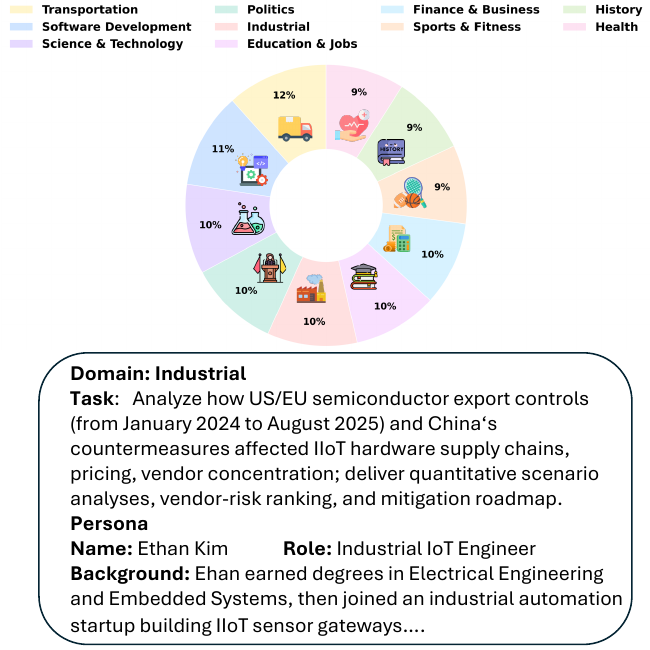}
    \vspace{-6mm}
    \caption{Domain Distribution and Example.}
    \label{fig:task stats}
    \vspace{-3mm}
\end{figure}

\subsection{Benchmark Tasks}
To mitigate evaluation costs, we curated 100 high-quality tasks based on human rankings (statistics and examples in Figure~\ref{fig:task stats}). 
This selection reflects practical constraints rather than pipeline deficiencies; as shown in Table~\ref{tab:expert_agreement}, the majority of automatically generated tasks satisfy deep research criteria. 
Thus, our pipeline enables the continuous generation of fresh, high-quality tasks, allowing our framework to function as a dynamic ``live'' benchmark suitable for long-term monitoring.

\section{Agentic Evaluation }
\label{sec:evalution}

\begin{figure*}[!t]
    \centering
    \includegraphics[width=0.96\linewidth]{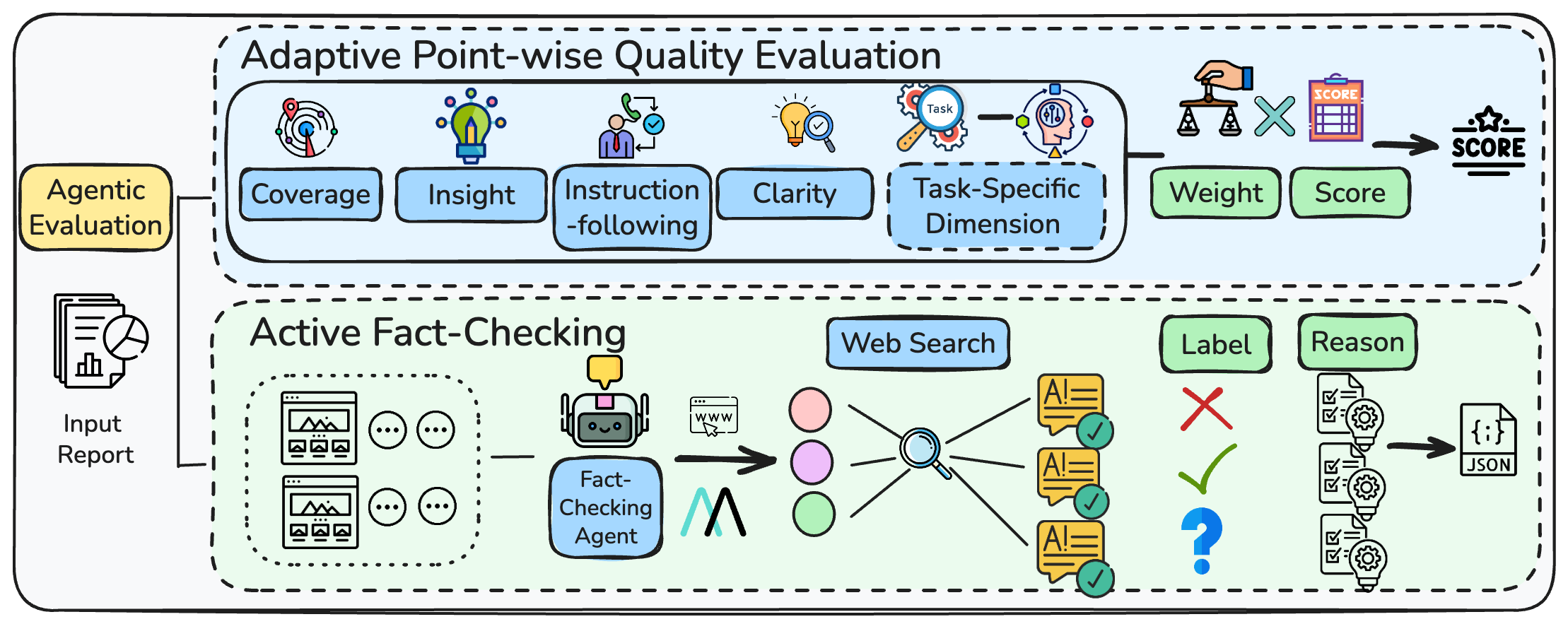}
    \vspace{-1ex}

    \caption{
    Overview of the proposed pipeline.
{(Top) Adaptive Point-wise Quality Evaluation} augments $\mathcal{D}_{\text{general}}$ with task-specific $\mathcal{D}_{\text{task}}$. An LLM scores criteria $s_{d,c}$, aggregating them into $S_{\text{quality}}$ via weights $W_d$ and $w_{d,c}$.
{(Bottom) Active Fact-Checking} extracts statements $\mathcal{S}_i$ from report segments $\{p_i\}$. An agent verifies claims using MCP-based retrieval, producing JSON labels (\texttt{Right}, \texttt{Wrong}, \texttt{Unknown}).
    }
    \label{fig:example}
\end{figure*}

In this section, we present our agentic evaluation pipeline for assessing deep research reports. As illustrated in Figure~\ref{fig:example}, the pipeline comprises two components: (1) an adaptive point-wise quality evaluator that actively derives task-specific evaluation dimensions, criteria, and relative weights conditioned on the given research task, enabling fine-grained and task-aware scoring, and (2) an active fact checker that verifies both cited and uncited statements through external evidence retrieval.

\subsection{Adaptive Point-wise Quality Evaluation}

The deep research system produces long-form reports that vary substantially across tasks and domains, making it insufficient to evaluate all outputs using a fixed and uniform rubric.
Prior work~\citep{du2025deepresearch,fan2025understanding} typically relies on a small set of pre-defined dimensions, which limits their ability to reflect task-specific evaluation aspects.
Meanwhile, manually constructing customized rubrics for each task~\citep{gou2025mind2web,yao2025rigorous,wang2025liveresearchbench} is labor-intensive and does not scale.

To address these challenges, as illustrated in Figure~\ref{fig:example}, we propose an adaptive point-wise quality evaluation framework.
For each task, the evaluator combines a fixed set of general dimensions with automatically generated task-specific dimensions, and assigns normalized weights to all dimensions to reflect their relative importance.
Each dimension is further instantiated with weighted evaluation criteria, enabling fine-grained, criterion-level scoring.
The final quality score is obtained by aggregating criterion scores within each dimension and then combining all dimensions according to their task-specific weights.

\textbf{Agentic Quality Evaluation Framework.} 
Formally, for a given task $t$, the evaluator first defines four general evaluation dimensions $\mathcal{D}_{\text{general}}$: \textit{Coverage}, \textit{Insight}, \textit{Instruction-following}, and \textit{Clarity}, capturing essential report qualities applicable across tasks. Dimension definitions are provided in Appendix~\ref{apd:evaluation quality}.
Then the evaluator generates a set of {task-specific} dimensions $\mathcal{D}_{\text{task}}$, tailored to task $t$.
For instance, in a task that compares policies across different countries and requires the specification of quantitative indicators, the task-specific dimensions include Metric Utility and Comparative Synthesis, which are important evaluation metrics in political analysis but may not apply to more general tasks (see Appendix~\ref{apd:sec example} for an example).
The full dimension set is: $\mathcal{D} = \mathcal{D}_{\text{general}} \cup \mathcal{D}_{\text{task}}$, 
The evaluator assigns a normalized weight $W_d$ to each dimension $d \in \mathcal{D}$ such that $ \sum_{d \in \mathcal{D}} W_d = 1$, where higher weights indicate greater importance of a dimension for evaluating the task.

For each dimension $d$, a set of criteria $\{c\}$ are generated along with their corresponding weights $w_{d,c}$, where $\sum_{c} w_{d,c} = 1$. 
Given a report $R$, the evaluator scores each criterion on a scale of $[1,10]$:
\begin{equation}
s_{d,c} = \mathrm{LLM}_{\theta}\!\left(R, c\right), \quad s_{d,c} \in [1, 10],
\label{eq:llm_scoring}
\end{equation}
The final evaluation score for task $t$ is computed as:
\begin{equation} 
S_{\text{quality}} = \sum_{d \in \mathcal{D}} W_d \sum_{c} w_{d,c} \, s_{d,c}.
\end{equation} 
The evaluator generates a task-specific overall quality score, offering greater relevance to the evaluation context. Furthermore, it enables granular analysis of individual dimensions and criteria, ensuring a more comprehensive understanding of report quality.
Details on prompts, LLM configurations, and full dimension set, dimension-level weights, criteria, and criterion-level weights can be found in Appendix~\ref{apd:evaluation quality}.

\subsection{Active Fact-Checking}
\label{sub sec: factual eval}

The adaptive point-wise quality evaluation provides fine-grained scoring over report quality dimensions.
However, it does not explicitly evaluate factual correctness, which is particularly critical for deep research reports.
Existing methods~\cite{du2025deepresearch,fan2025understanding} typically check if citations support the text, but this paradigm fails when: i) reports lack citations; ii) claims appear in uncited segments; and (iii) citation-based verification checks whether a cited source supports a claim, rather than its factual correctness. 
To address these issues, we propose an active fact checking framework that 
actively retrieves and examines external evidence to assess the factual consistency of the 
entire report, as shown in Figure~\ref{fig:example}.

\textbf{Agentic Fact-Checking Framework.}
Built upon \textit{MiroFlow}~\citep{2025miroflow}, our agent iteratively invokes MCP tools to retrieve external evidence. Rather than relying solely on citations, it proactively identifies and verifies claims to ensure comprehensive statement-level checking.

Given a generated report that requires factual evaluation, our fact-checking agent follows a structured, multi-stage pipeline to support fine-grained, statement-level verification of long-form reports.
To reduce the challenges associated with long-context processing in lengthy reports and to enable parallel verification across segments, the agent first segments $R$ into a set of smaller parts $R \;\rightarrow\; \mathcal{P} = \{p_1, p_2, \dots, p_N\}$,
For each input part $p_i$, the agent extracts a set of statements $\mathcal{S}_i = \{ s_{i1}, s_{i2}, \dots \}$ involving verifiable entities such as
\textit{numbers, news, events, dates, locations, or people}.

For each statement $s \in \mathcal{S}_i$, the agent invokes a retrieval tool to search the web and collect relevant evidence $\mathcal{E}(s)$.
Although verification is performed at the statement level, the agent holds the {full segment context $p_i$ as well as the associated deep research task}, enabling context-aware and task-consistent judgments.
Based on the consistency between $s$ and the retrieved evidence, the agent assigns one of three labels: $y(s) \in \{\texttt{Right},\; \texttt{Wrong},\; \texttt{Unknown}\}$.
\texttt{Right} denotes support, \texttt{Wrong} indicates contradiction, and \texttt{Unknown} marks insufficient evidence, explicitly distinguishing unverifiable claims from errors.
 
Results, including labels, evidence, and reasoning, are returned in JSON format. Implementation details and examples are in Appendix~\ref{apd:evaluation fact} and~\ref{apd:sec example}. Finally, the \texttt{Ratio} metric of factual evaluation is defined as the proportion of right statements over all statements: $\texttt{Ratio} = \frac{N_{\texttt{Right}}}{N_{\texttt{Statements}}}$.

\section{Experiments}
\subsection{Experimental Setup}
\begin{table*}[!htbp]
\small
\centering
\caption{\textbf{Quality evaluation results across different deep research system.}
Bold numbers indicate the best scores.}
\vspace{-2mm}
\resizebox{0.86\linewidth}{!}{%
\begin{tabular}{l>{\columncolor{avgcol}}c ccccc}
\toprule
\textbf{Model} & \textbf{Avg} & \textbf{Covera.} & \textbf{Insight} & \textbf{Instr.} & \textbf{Clarity} & \textbf{Task.} \\
\midrule

\rowcolor{avg5}
DeepSeek Deep Research & \mixgray{avg5}{5.25} & 5.9 & 5.2 & 7.2 & 8.4 & 4.3 \\

\rowcolor{avg5}
Manus & \mixgray{avg5}{5.95} & 7.2 & 5.8 & 8.3 & 7.1 & 5.2 \\

\rowcolor{avg6}
Perplexity Deep Research & \mixgray{avg6}{6.86} & 8.2 & 6.6 & 9.3 & 8.6 & 5.9 \\

\rowcolor{avg6}
Grok4 Deep Research & \mixgray{avg6}{6.92} & 8.5 & 6.6 & 9.6 & 8.2 & 6.0 \\

\rowcolor{avg7}
Doubao Deep Research& \mixgray{avg7}{7.06} & 8.6 & 7.0 & 9.2 & 7.7 & 6.3 \\

\rowcolor{avg7}
Qwen-3-235B-A22B Deep Research & \mixgray{avg7}{7.17} & 8.0 & 7.9 & 8.7 & 8.3 & 6.6 \\

\rowcolor{avg7}
OpenAI Deep Research & \mixgray{avg7}{7.28} & 8.6 & 7.3 & 9.0 & 7.6 & 6.7 \\

\rowcolor{avg7}
Claude-Sonnet-4.5 Deep Research & \mixgray{avg7}{7.53} & 8.8 & 8.0 & 9.2 & 7.8 & 6.8 \\

\rowcolor{avg8}
Gemini-2.5-Pro Deep Research & \mixgray{avg8}{\textbf{8.51}} & \textbf{9.2} & \textbf{9.0} & \textbf{9.7} & \textbf{9.1} & \textbf{8.0} \\

\bottomrule
\end{tabular}
}
\label{tab:model_comparison}
\end{table*}
We evaluate 9 major commercial deep research systems, including OpenAI Deep Research~\citep{openai2025deepresearch}, Gemini-2.5-Pro Deep Research~\citep{google2025deepresearch}, Grok4 Deep Research~\citep{xai2025deepsearch}, Claude-Sonnet-4.5 Deep Research~\citep{anthropic2025claude45}, Qwen3-235B-A22B Deep Research~\citep{yang2025qwen3}, DeepSeek Deep Research~\citep{liu2024deepseek}, Perplexity Deep Research~\citep{perplexity2025deepresearch}, Doubao Deep Research~\citep{doubao2025deepresearch}, and Manus~\citep{manus2025}.
For each deep research system, we collect \textbf{100 reports} by running the system on the deep research tasks constructed in our pipeline~\ref{sec:pipeline}. Details about the collection of deep research systems are provided in Appendix~\ref{apd:DRAs}.

Following Sec.~\ref{sec:evalution}, we utilize \textit{Gemini-2.5-pro}~\cite{comanici2025gemini25pushingfrontier} for \textit{Adaptive Point-Wise Quality Evaluation} to generate all adaptive components (dimensions, weights, criteria) and produce final scores. For active fact checking, we implement the agent on \textit{MiroFlow}~\citep{2025miroflow} using \textit{GPT-5-mini}~\cite{openai2025gpt5systemcard} (default settings). The agent employs \textit{Google Serper API} for retrieval, with a maximum of 30 agent turns. See Appendix~\ref{apd:evaluation} for details.

\subsection{Main Results}

\textbf{Overall Quality Evaluation.}
Table~\ref{tab:model_comparison} presents a point-wise quality evaluation across nine representative deep research systems.
We observe clear stratification: Gemini-2.5-Pro Deep Research achieves the highest average score ($8.51$) and leads across all dimensions, followed by Claude-Sonnet-4.5 Deep Research ($7.53$).
This advantage is driven by \textit{Coverage}, \textit{Insight}, and \textit{Instruction-following}, where these top systems exceed $8.5$, indicating strong abilities in information gathering, synthesis, and execution of complex instructions.

DeepSeek and Manus show moderate \textit{Instruction-following} ($7.2$, $8.3$) but lag in \textit{Coverage} and \textit{Insight}, resulting in lower overall scores ($5.25$, $5.95$).
Perplexity and Grok4 improve substantially in \textit{Coverage} ($>8.2$) and \textit{Instruction-following} ($9.3$, $9.6$), reflecting stronger retrieval and planning.
Doubao and Qwen-3-235B-A22B further enhance analytical depth, with Qwen achieving a high \textit{Insight} score of $7.9$.
Finally, OpenAI and Claude-Sonnet-4.5 exhibit balanced performance across dimensions.

Notably, task-specific scores are consistently lower than general scores across all systems. This indicates that while systems excel at general synthesis, they often fail to optimize for task-specific criteria. This gap motivates our adaptive dimensions, which capture quality aspects missed by fixed rubrics. Ultimately, generating high-quality task-specific content remains a key challenge for current deep research systems.

\begin{table*}[h]
\small
\centering
\caption{\textbf{Factual evaluation results across different deep research system.} 
Bold numbers indicate the best values.}
\vspace{-2mm}
\resizebox{0.88\linewidth}{!}{%
\begin{tabular}{l>{\columncolor{avgcol}}ccccc}
\toprule
\textbf{Model} & \textbf{Ratio} & \textbf{Statements} & \textbf{Right} & \textbf{Wrong} & \textbf{Unknown} \\
\midrule
\rowcolor{avg5}
Perplexity Deep Research & \mixgray{avg5}58.94\% & 61.34 & 36.16 & 9.08 & 16.10 \\
\rowcolor{avg6}
Claude-Sonnet-4.5 Deep Research& \mixgray{avg6}60.72\% & 57.30 & 34.79 & 6.16 & 16.35 \\
\rowcolor{avg6}
Grok4 Deep Research& \mixgray{avg6}61.81\% & 47.16 & 29.15 & 5.44 & 12.57 \\
\rowcolor{avg6}
Doubao Deep Research& \mixgray{avg6}69.50\% & 80.75 & 56.12 & 7.43 & 17.20 \\
\rowcolor{avg7}
Qwen-3-235B-A22B Deep Research & \mixgray{avg7}72.39\% & 37.45 & 27.11 & 3.36 & 6.34 \\
\rowcolor{avg7}
OpenAI Deep Research & \mixgray{avg7}76.21\% & 45.98 & 35.04 & 2.72 & 8.22 \\
\rowcolor{avg7}
DeepSeek Deep Research& \mixgray{avg7}76.44\% & 25.08 & 19.17 & 1.81 & 4.10 \\
\rowcolor{avg7}
Gemini-2.5-Pro Deep Research & \mixgray{avg7}76.62\% & \textbf{86.99} & 66.65 & 4.16 & 16.18 \\
\rowcolor{avg8}
Manus & \mixgray{avg8}\textbf{82.30\%} & 57.90 & 47.65 & 2.23 & 8.02 \\
\bottomrule
\end{tabular}
}
\label{tab:factual_results}
\end{table*}

\textbf{Factual Evaluation.}
Table~\ref{tab:factual_results} presents factual evaluation results where our agent assesses statements per report. We report average checkable \texttt{Statements} and counts for \texttt{Right}, \texttt{Wrong}, and \texttt{Unknown} claims (defined in Sec.~\ref{sub sec: factual eval}). Models are ranked in ascending order by factual \texttt{Ratio}.

Top performers like Manus, Gemini-2.5-Pro, and DeepSeek achieve ratios $>76\%$, indicating superior reliability.
In contrast, Perplexity and Claude-Sonnet-4.5 exhibit lower ratios, implying more unverifiable or incorrect statements.
Substantial variation exists in statement volume: Gemini-2.5-Pro and Doubao produce notably more claims ($86.99$, $80.75$), yielding denser reports, whereas DeepSeek adopts a conservative strategy ($25.08$).
These results suggest a potential trade-off between maintaining high factual accuracy and increasing the volume of reported statements.

Systems with higher \texttt{Ratio}, such as Manus and DeepSeek, exhibit consistently low \texttt{Wrong} counts ($2.23$, $1.81$), indicating strong avoidance of false claims.
Conversely, lower \texttt{Ratio} systems show higher \texttt{Unknown} values, implying claims are often unsupported rather than explicitly incorrect.
Notably, \texttt{Wrong} statements are rare compared to \texttt{Unknown} across all systems, suggesting factual risks stem more from weakly grounded claims than outright errors.

\subsection{Validation of Evaluation Methods}
\label{subsec: validation of eval}

To validate the reliability of methods, we conduct an analysis for three
dimensions: cross-judge consistency, stochastic stability, human-model alignment.

\textbf{Cross-judge Consistency of Quality Evaluation.} 
To mitigate self-preference bias in Gemini-2.5-Pro (primary judge), we employ GPT-5~\cite{openai2025gpt5systemcard} as a secondary judge.
Although GPT-5 is stricter (lower scores; see Table~\ref{tab:model_comparison_gpt5}), rankings remain highly consistent with Table~\ref{tab:model_comparison}: 7 of 9 models hold identical positions.
Only Doubao and Qwen exhibit a minor swap ($|\Delta\text{Rank}| = 1$), suggesting that the overall ranking is highly robust.
In addition, the Task-Specific dimension is consistently identified by the GPT-5 judge as the lowest-scoring aspect, highlighting both its importance and its inherent difficulty.

\begin{table}[t]
\centering
\caption{{Quality evaluation results using GPT-5 judge.}
The last column reports the absolute rank difference compared to Table~\ref{tab:model_comparison}.}
\vspace{-2mm}
\resizebox{0.9\linewidth}{!}{%
\begin{tabular}{l>{\columncolor{avgcol}}c c}
\toprule
\textbf{Model} & \textbf{Avg} & $\lvert\Delta\textbf{Rank}\rvert$ \\
\midrule
DeepSeek Deep Research & 2.72 & 0 \\
Manus & 3.60 & 0 \\
Perplexity Deep Research & 4.08 & 0 \\
Grok4 Deep Research & 4.18 & 0 \\
Qwen-3-235B-A22B Deep Research & 4.23 & 1 \\
Doubao Deep Research & 4.46 & 1 \\
OpenAI Deep Research & 4.63 & 0 \\
Claude-Sonnet-4.5 Deep Research & 4.73 & 0 \\
Gemini-2.5-Pro Deep Research & \textbf{5.29} & 0 \\
\bottomrule
\end{tabular}
}
\label{tab:model_comparison_gpt5}
\end{table}

\textbf{Stochastic Stability of Quality Evaluation.}
We assess stochastic stability via three independent runs using Gemini-2.5-Pro. As shown in Table~\ref{tab:stability}, rankings remain unchanged with minimal score standard deviations, demonstrating the evaluation's high stability against randomness.

\begin{table}[t]
\centering
\caption{Quality evaluation across 3 independent runs.}
\vspace{-2mm}
\label{tab:stability}
 \resizebox{0.98\linewidth}{!}{%
\begin{tabular}{lcc}
\toprule
\textbf{Model} & \textbf{Score ($\mu \pm \sigma$)} & \textbf{Rank} \\
\midrule
DeepSeek Deep Research & 5.24 ($\pm 0.02$) & 9.0 \\
Manus & 5.92 ($\pm 0.02$) & 8.0 \\
Perplexity Deep Research & 6.85 ($\pm 0.01$) & 7.0 \\
Grok4 Deep Research & 6.95 ($\pm 0.04$) & 6.0 \\
Doubao Deep Research & 7.08 ($\pm 0.02$) & 5.0 \\
Qwen-3-235B-A22B Deep Research & 7.21 ($\pm 0.06$) & 4.0 \\
OpenAI Deep Research & 7.30 ($\pm 0.08$) & 3.0 \\
Claude-Sonnet-4.5 Deep Research & 7.51 ($\pm 0.01$) & 2.0 \\
Gemini-2.5-Pro Deep Research & 8.52 ($\pm 0.03$) & 1.0 \\ \bottomrule
\end{tabular}
}
\end{table}

\begin{figure}[t]
    \centering
    \includegraphics[width=0.7\linewidth]{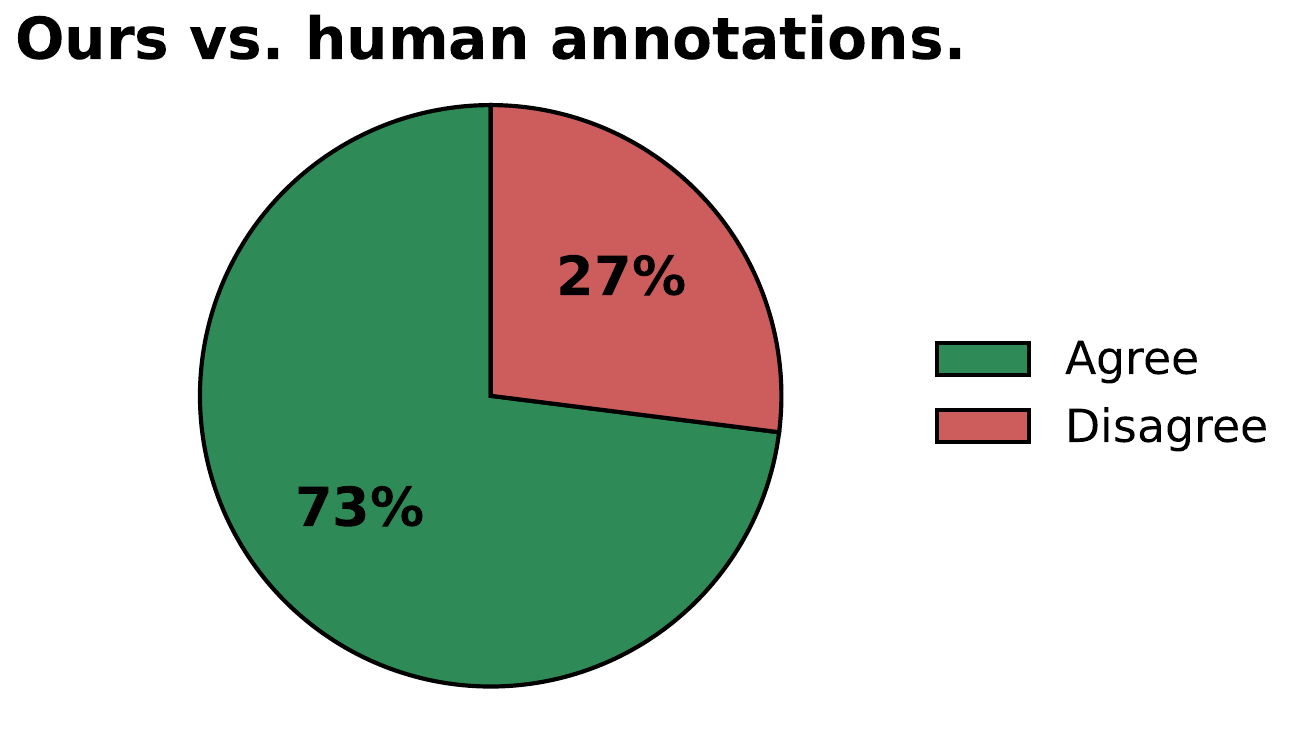}
    \caption{{Agreement between our annotations and human experts.}}
    \label{fig:ours_vs_human}
\end{figure}

\textbf{Human--Model Alignment.}
To validate our \emph{active fact-checking} module, four experts annotated $80$ statements.
Treating both \emph{Wrong} and \emph{Unknown} as negative. As shown in Figure~\ref{fig:ours_vs_human}, we achieve $73\%$ agreement .
This suggests the agent approaches human performance, aligning with prior work~\cite{wei2024long}.

We manually re-annotated the 20 inconsistent statements using a human expert assisted by GPT-5.2.
Analysis reveals the automated evaluation was correct in $70\%$ of cases (vs. $30\%$ for humans), primarily due to its exhaustive verification capabilities.
Examples of correct automated judgments and failure cases are provided in Appendix~\ref{apd:example agent right} and~\ref{apd:example agent wrong}, respectively.

\section{Conclusions}
In this paper, we introduce an automated framework for deep research task construction and
agentic evaluation in report quality and factuality. Our persona-driven task construction enables generation of realistic, complex tasks without manual annotation. We propose an adaptive point-wise quality evaluation for report assessment, together with an active fact-checking via external evidence retrieval. 
Experiments on nine deep research systems reveal substantial performance differences, demonstrating the effectiveness of our framework in evaluating generated long report.

\section*{Limitations}

Despite its effectiveness, the proposed framework has several practical limitations. The current implementation is largely English-centric: although the persona-driven task construction and adaptive evaluation mechanisms are language-agnostic, the benchmark tasks, evidence sources, and reporting pipelines are grounded in English-speaking information ecosystems. As a result, performance in multilingual settings and the ability to synthesize evidence across diverse languages remain unexplored.

In addition, the agentic evaluation pipeline incurs substantial computational and financial costs. The framework relies on frequent interactions with frontier models, using Gemini-2.5-Pro for quality scoring and GPT-5-mini for factual verification, alongside extensive Google Serper API usage. While the fact checking agent’s multi turn, tool intensive design enables high evaluation depth, it constrains scalability for large scale or real time deployment under limited resources.

\bibliography{iclr2026/acl}

\newpage
\appendix
\section{Usage of AI Assistant}

We use ChatGPT solely for language refinement of the manuscripts text. All conceptual content, experimental design, analysis, and conclusions are developed entirely by the authors. We carefully review the AI-assisted edits to ensure that the meaning and technical accuracy of the original text are fully preserved.

\section{Deep Research Systems Details}
\label{apd:DRAs}

Table~\ref{tab:data_collection_time} reports the main time windows during which reports from different deep research systems were collected. All dates correspond to the year 2025. Reports from the other 9 deep research systems were generated and downloaded from their official websites using automated tools.

\begin{table*}[h]
\centering
\caption{\textbf{Primary data collection periods of deep research systems (2025).}}
\label{tab:data_collection_time}
\begin{tabular}{lll}
\toprule
\textbf{Deep Research System} & \textbf{Data Collection Date Range} & \textbf{Avg Length} \\
\midrule
Claude-Sonnet-4.5 Deep Research  & Aug 19 -- Aug 28                     & 26.3K \\
Doubao Deep Research             & Aug 19 -- Aug 26; Sep 1 -- Sep 7     & 48.4K \\
Gemini-2.5-Pro Deep Research     & Aug 19 -- Aug 26; Sep 5 -- Sep 6     & 51.8K \\
Perplexity Deep Research         & Aug 22 -- Aug 26                     & 13.7K \\
OpenAI Deep Research             & Aug 27 -- Sep 8                      & 41.3K \\
Grok4 Deep Research              & Aug 28 -- Sep 1                      & 11.0K \\
Manus                            & Aug 28 -- Sep 8                      & 30.8K \\
Qwen3-235B-A22B Deep Research    & Aug 29                               & 29.8K \\
DeepSeek Deep Research           & Nov 10                               & 5.5K \\
\bottomrule
\end{tabular}
\end{table*}

Avg Length denotes the average length of valid Deep Research outputs produced by each deep research system across all evaluated tasks. Most Deep Research Agents produce responses exceeding ten thousand characters on average. In particular, Gemini-2.5-Pro, Doubao, and OpenAI Deep Research generate substantially longer outputs, with average lengths reaching several tens of thousands of characters.

\section{More Results}

Table~\ref{tab:model_comparison_gpt5_more} presents the results of adaptive point-wise quality evaluation using GPT-5 judge.

\begin{table*}[h]
\centering
\caption{\textbf{Adaptive point-wise quality evaluation  full results using GPT-5 judge.}
The last column reports the absolute rank difference compared to Table~\ref{tab:model_comparison}.}
\resizebox{1.0\linewidth}{!}{%
\begin{tabular}{l>{\columncolor{avgcol}}c ccccc c}
\toprule
\textbf{Model} & \textbf{Avg} & \textbf{Covera.} & \textbf{Insight} & \textbf{Instr.} & \textbf{Clarity} & \textbf{Task.} & $\lvert\Delta\textbf{Rank}\rvert$ \\
\midrule

DeepSeek Deep Research & 2.72 & 3.5 & 3.5 & 5.0 & 4.0 & 1.8 & 0 \\

Manus & 3.60 & 5.1 & 4.3 & 6.4 & 5.0 & 2.3 & 0 \\

Perplexity Deep Research & 4.08 & 5.6 & 4.8 & 6.9 & 5.4 & 2.8 & 0 \\

Grok4 Deep Research & 4.18 & 5.9 & 4.8 & 7.5 & 5.3 & 2.8 & 0 \\

Qwen-3-235B-A22B Deep Research & 4.23 & 6.0 & 5.9 & 6.5 & 3.2 & 3.0 & 1 \\

Doubao Deep Research & 4.46 & 6.4 & 5.4 & 7.2 & 5.2 & 3.1 & 1 \\

OpenAI Deep Research & 4.63 & 6.6 & 5.9 & 7.3 & 5.0 & 3.2 & 0 \\

Claude-Sonnet-4.5 Deep Research & 4.73 & 6.6 & 6.0 & 7.0 & 4.8 & 3.4 & 0 \\

Gemini-2.5-Pro Deep Research & \textbf{5.29} & \textbf{7.0} & \textbf{7.1} & \textbf{7.9} & \textbf{6.4} & \textbf{3.7} & 0 \\

\bottomrule
\end{tabular}
}
\label{tab:model_comparison_gpt5_more}
\end{table*}

\section{Evaluation Methods Details}
\label{apd:evaluation}

\subsection{Adaptive Point-wise Quality Evaluation}
\label{apd:evaluation quality}

For the \textbf{adaptive point-wise quality evaluation}, we define four general evaluation dimensions.
\textit{Coverage}: Breadth, depth, and relevance of coverage. \textit{Insight}:  Depth, originality, logic, and value of analysis. \textit{Instruction-following}: Accuracy in meeting all requirements and constraints. \textit{Clarity}: Readability, fluency, structure, and ease of understanding. 
In addition, the framework automatically generates between one and three \textit{task-specific} dimensions. 
For each dimension, we create between one and ten criteria, each scored on a $[0,10]$ scale with two decimal places of precision.

For the base LLM, we employ \textbf{Gemini-2.5-Pro} to generate the dimensions, criteria, and weights, 
using a maximum of 8192 new tokens, a temperature of 0.1, and a random seed of 42. 
We additionally use \textbf{GPT-5} for scoring in Sec~\ref{subsec: validation of eval}, with a maximum of 8192 new tokens 
and its default temperature setting.

\textbf{Adaptive Point-wise Quality Evaluation Prompt.} We present the prompts used for task-specific dimension generation, followed by the prompts for assigning weights to the four fixed dimensions and the additional task-specific dimensions, as well as the prompts for generating evaluation criteria and corresponding weights for each dimension. Finally, we provide the prompt used to perform scoring with an LLM-based judge. All prompts are designed to return outputs in a JSON format.

\begin{tcblisting}{title={Point-wise Task-Specific Dimension Generation},colback=lightgrey,colframe=black,arc=1mm,boxrule=1pt,left=1mm,right=1mm,top=1mm,bottom=1mm,breakable,fontupper=\tiny\ttfamily,listing only,listing engine=listings,listing options={breaklines,breakautoindent=false,breakindent=0pt,keepspaces,tabsize=4,literate={“}{"}{1} {”}{"}{1} {‘}{'}{1} {’}{'}{1} {—}{--}{1} {–}{-}{1} { }{ }{1}
    }}
<system_role>

You are an expert evaluator who designs **query-specific meta-evaluation dimensions** for deep research reports. Your goal is to identify unique quality aspects that matter for a given task, beyond the four standard meta-dimensions.
</system_role>

<user_prompt>
**Standard Meta-Dimensions** (already covered):
1. **Coverage**: Breadth, depth, and relevance of coverage
2. **Insight**: Depth, originality, logic, and value of analysis
3. **Instruction Following**: Accuracy in meeting all requirements
4. **Clarity**: Clarity, fluency, structure, and ease of understanding

**Your Task**: For the research task below, generate **1–3 additional same-level meta-evaluation dimensions** that are:
- Highly specific to this query
- Distinct from the four standard meta-dimensions
- Crucial for assessing quality in this domain
- Actionable and measurable
- Serve as **upper-level meta-dimensions** that can be further expanded into more detailed evaluation standards 
- Do NOT include any factuality-related meta-dimensions, since factual accuracy is handled by a separate evaluation system

<research_task>
"{task_prompt}"
</research_task>

**Guidelines**:
1. Analyze the task carefully to understand its domain, methodology, data needs, and unique challenges.
2. Identify domain-specific quality factors (e.g., for finance: market timing; for science: experimental validity; for policy: stakeholder impact).
3. Create meta-dimensions that are:
   - Unique to this query type (not generic)
   - Non-overlapping with the four standard meta-dimensions
   - Focused on specialized aspects relevant to this domain
4. For each meta-dimension, provide:
   - **Name** (1–3 words)
   - **Definition** (short explanation of what it measures and why it matters)

**Output Format**:
Return only a JSON list of meta-dimensions, e.g.:

<json_output>
[
  {{
    "meta_dimension_name": "Xxx",
    "definition": "Clear, concise explanation"
  }},
  ...
]
</json_output>

</user_prompt>

\end{tcblisting}
\begin{tcblisting}{title={Point-wise Weight Generation},colback=lightgrey,colframe=black,arc=1mm,boxrule=1pt,left=1mm,right=1mm,top=1mm,bottom=1mm,breakable,fontupper=\tiny\ttfamily,listing only,listing engine=listings,listing options={breaklines,breakautoindent=false,breakindent=0pt,keepspaces,tabsize=4,literate=
            {“}{"}{1}       % 左弯引号 -> 直引号
            {”}{"}{1}       % 右弯引号 -> 直引号 (修复 Byte 93)
            {‘}{'}{1}       % 左弯撇 -> 直撇
            {’}{'}{1}       % 右弯撇 -> 直撇
            {—}{--}{1}      % 破折号 -> 两个减号
            {–}{-}{1}       % En-dash (1–3中间那个) -> 减号
            {±}{+/-}{2}     % 正负号 (修复 Byte B1) -> 替换为 +/-
            { }{ }{1}       % 不间断空格 -> 普通空格
    }}
<system_role>
You are a senior research evaluation expert. Your job is to (1) consider both the four fixed meta-dimensions and the provided query-specific meta-dimensions (each with a name and definition), and (2) assign **dynamic, well-justified weights** to all dimensions so that the total equals **1.0**.
</system_role>

<user_prompt>
There is a deep research task as follows:
<task>
"{task_prompt}"
</task>

**Fixed Meta-Dimensions (always included):**
[
  {{
    "meta_dimension_name": "Coverage",
    "definition": "Breadth, depth, and relevance of coverage."
  }},
  {{
    "meta_dimension_name": "Insight",
    "definition": "Depth, originality, logic, and value of analysis."
  }},
  {{    
    "meta_dimension_name": "Instruction Following",
    "definition": "Accuracy in meeting all requirements and constraints."
  }},
  {{
    "meta_dimension_name": "Clarity",
    "definition": "Readability, fluency, structure, and ease of understanding."
  }}
]

**Provided Query-Specific Meta-Dimensions (each includes name + definition)**
<additional_meta_dimensions_json>
{additional_dimensions_json}
</additional_meta_dimensions_json>

**Your Goals**
1. **Task-grounded analysis**: Carefully analyze the <task> to identify goals, constraints, risks, and success criteria.
2. **Dynamic weighting**: Assign a weight (0–1) to **each** dimension (fixed + provided).  
   - The **sum across all dimensions must be exactly 1.0**.
   - Weights should reflect the **unique characteristics** of the <task> (do not use fixed presets).
3. **Specific justification**: In <analysis>, explicitly justify **each** weight by referencing the <task> and, for the provided meta-dimensions, their **definitions**. Avoid generic statements.

**Constraints & Notes**
- Do **not** introduce new dimensions. Only use the fixed four plus the provided query-specific ones.
- Do **not** include any factuality-related dimensions (factuality is evaluated elsewhere).
- Avoid overlap: if a provided dimension substantially overlaps with a fixed one, explain how you differentiate it in scope and why its weight is still necessary.
- If no additional dimensions are provided (empty list), distribute weights among the four fixed dimensions only.

**Output Format (STRICT)**
First produce a concise yet concrete <analysis> that:
- Explains the task-grounded reasoning behind the overall weighting strategy.
- Gives a 1–3 sentence justification **for each dimension** tying the weight to the task and (for provided ones) their definitions.

Then produce <json_output> containing only the final weights, with keys exactly matching the dimension names:
- Fixed keys (always present): "comprehensiveness", "insight", "instruction_following", "readability"
- For each provided meta-dimension, use its exact "meta_dimension_name" as the key.

Example shape:
<analysis>
(Your reasoning here. One short paragraph about overall trade-offs, then bullet points or short lines justifying each dimension's weight.)
</analysis>

<json_output>
{{
  "coverage": 0.xx,
  "insight": 0.xx,
  "instruction_following": 0.xx,
  "clarity": 0.xx,
  "additional_dimension": 0.xx
}}
</json_output>

**Validation**
- Ensure all weights are in [0, 1].
- Ensure the sum across all keys equals **1.00** (allowing up to ±0.001 rounding; otherwise adjust and state the adjustment briefly in <analysis>).
- Do not output anything other than <analysis> and <json_output>.
</user_prompt>

\end{tcblisting}
\begin{tcblisting}{title={Point-wise Criteria Generation},colback=lightgrey,colframe=black,arc=1mm,boxrule=1pt,left=1mm,right=1mm,top=1mm,bottom=1mm,breakable,fontupper=\tiny\ttfamily,listing only,listing engine=listings,listing options={breaklines,breakautoindent=false,breakindent=0pt,keepspaces,tabsize=4,literate={“}{"}{1} {”}{"}{1} {‘}{'}{1} {’}{'}{1} {—}{--}{1} {–}{-}{1} { }{ }{1}
    }}
<system_role>
You are an expert evaluator of research reports. Your job is to break down an meta-evaluation dimension into clear, specific, task-relevant criteria with explanations and weights.
</system_role>

<user_prompt>
We evaluate a research report written for the task below across {num_dimensions} meta evaluation dimensions:
{meta_dimensions}

<task>
"{task_prompt}"
</task>

<instruction>
Your goal: For the **{dimension_name}** dimension, generate task-specific evaluation criteria.

Steps:
1. **Analyze Task**: Identify the essential areas and coverage needed to satisfy "{dimension_name}".
2. **Formulate Criteria**: Write diverse, non-overlapping criteria items.
3. **Explain Rationale**: Provide a short explanation (`explanation`) for each criterion.
4. **Assign Weights**: Give each criterion a weight (`weight`) so that the total = **1.0**. Adjust the last item if needed.
5. **Focus**: Stay strictly within "{dimension_name}", avoiding overlap with the other dimensions.

Output format:
1. First, provide `<analysis>` explaining your reasoning and weight allocation.
2. Then output `<json_output>` as a list of criteria in the format:

<json_output>
[
  {{
    "criterion": "...",
    "explanation": "...",
    "weight": ...
  }},
  ...
]
</json_output>

Now begin for the task above and dimension = **{dimension_name}**.
</user_prompt>

\end{tcblisting}
\begin{tcblisting}{title={Point-wise Score Prompt},colback=lightgrey,colframe=black,arc=1mm,boxrule=1pt,left=1mm,right=1mm,top=1mm,bottom=1mm,breakable,fontupper=\tiny\ttfamily,listing only,listing engine=listings,listing options={breaklines,breakautoindent=false,breakindent=0pt,keepspaces,tabsize=4,literate={“}{"}{1} {”}{"}{1} {‘}{'}{1} {’}{'}{1} {—}{--}{1} {–}{-}{1} { }{ }{1}
    }}
<system_role>
You are a strict, meticulous, and objective evaluator of deep research reports. 
You score the report on a **single evaluation dimension** at a time.  
You must evaluate strictly according to the provided **criteria under that dimension**. 
Do **not** evaluate factual accuracy (handled by a separate system).
</system_role>

<user_prompt>
**Task**
<task>
{task_prompt}
</task>

**Report to Evaluate**
<Report>
{report}
</Report>

**Evaluation Dimension and Criteria**
Below is the dimension to evaluate. It contains multiple criteria, each with a short explanation.  
<criteria_of_one_dimension_json>
{criteria_of_one_dimension_json}
</criteria_of_one_dimension_json>

**Scoring Rules**
- For **each criterion**, assign a **continuous score from 0 to 10** (real number), and provide a concise justification (`analysis`) grounded in the report content.
- **Revised Scale (more discriminative):**
  - **0–2 (Very Poor):** Severe deficiencies; almost entirely fails the criterion.
  - **2–4 (Poor):** Major gaps or flaws; only marginally addresses the criterion.
  - **4–6 (Fair):** Covers the basics but shallow, inconsistent, or with notable weaknesses.
  - **6–7.5 (Good):** Solid attempt; generally clear and adequate, but lacks depth or polish.
  - **7.5–9 (Very Good):** Strong and well-executed with only minor issues; above average.
  - **9–10 (Excellent):** Outstanding; fully satisfies and exceeds expectations in depth, clarity, and execution. Reserved for exceptional cases.
- Scores should reflect only the current criterion, avoiding overlap with other dimensions.
- **Do not** judge factual correctness; only judge coverage, reasoning quality, clarity, structure, domain-appropriateness, etc.
- Be conservative: most typical reports should fall in the **4–8 range**. Scores above 9 should be rare.

**Output Format (STRICT)**
Output `<json_output>` as a list of criteria in a valid JSON object format:


<json_output>
[
    {{
        "criterion": "text of the criterion",
        "analysis": "your justification",
        "report_score_0_to_10": x.xx
    }},
    {{
        "criterion": "text of the criterion",
        "analysis": "your justification",
        "report_score_0_to_10": x.xx
    }},
    ...
]
</json_output>

**Validation**
- Use the exact criterion names as keys in the JSON.
- Each score must be a real number in [0,10], rounded to two decimals.
- Ensure the JSON is strictly valid and parseable (no trailing commas, properly escaped characters).
</user_prompt>

\end{tcblisting}

\subsection{Active Fact-checking}
\label{apd:evaluation fact}

For the base LLM, we employ GPT-5-mini to segment the reports and perform fact-checking. 
The model is used with default temperature settings, a maximum of 128K new tokens, a 
190K maximum context length, top-$p$ = 1.0, and top-$k$ = $-1$. 

For retrieval, we use the Google Serper API to conduct searches and scrape website content. 
The number of returned results is set to 10, with a maximum of 5 retry attempts. 
For our agent workflow, we set the maximum number of turns to 30, with up to 10 tool calls allowed per turn. 
The agent interacts with the MCP server \texttt{tool-search}. 
The MCP tools available on the \texttt{tool-search} server mainly include \texttt{google\_search()} for search, 
\texttt{scrape\_website()} for retrieving webpage content, and 
\texttt{wiki\_get\_page\_content()} for obtaining content from specific Wikipedia pages corresponding to target entities.

\textbf{Active Fact-checking Prompt.} We present the prompt used to segment a Deep Research report into multiple paragraphs, after which each paragraph is independently evaluated by a fact-checking agent. We further show the agent system prompt used in the MiroFlow framework, along with the task description prompt for fact-checking. Finally, a summary prompt aggregates the evaluations into a structured JSON-formatted result.

\begin{tcblisting}{title={Factual Evaluation Report Segmentation},colback=lightgrey,colframe=black,arc=1mm,boxrule=1pt,left=1mm,right=1mm,top=1mm,bottom=1mm,breakable,fontupper=\tiny\ttfamily,listing only,listing engine=listings,listing options={breaklines,breakautoindent=false,breakindent=0pt,keepspaces,tabsize=4,literate={“}{"}{1} {”}{"}{1} {‘}{'}{1} {’}{'}{1} {—}{--}{1} {–}{-}{1} { }{ }{1}
    }}
You are a text segmentation assistant. Your task is to segment reports into logical parts:
- Avoid single-sentence parts.
- Do NOT alter the original content.
- Titles, headings, or section headers should be combined with the following content as one part.
- If multiple sentences clearly belong to the same section, keep them together in one part instead of splitting them.
- Output must be ONLY a valid JSON list, where each element is one part.

Example output format:
[
  "This is the first part.",
  "This is the second part.",
  "This is the third part."
]

Please segment the following report into logical parts:

{Deep Research Report}

\end{tcblisting}
\begin{tcblisting}{title={Agent System Prompt},colback=lightgrey,colframe=black,arc=1mm,boxrule=1pt,left=1mm,right=1mm,top=1mm,bottom=1mm,breakable,fontupper=\tiny\ttfamily,listing only,listing engine=listings,listing options={breaklines,breakautoindent=false,breakindent=0pt,keepspaces,tabsize=4,literate={“}{"}{1} {”}{"}{1} {‘}{'}{1} {’}{'}{1} {—}{--}{1} {–}{-}{1} { }{ }{1}
    }}
In this environment you have access to a set of tools you can use to answer the user's question.
Today is: {formatted_date}
Important Notes:
- Tool-use must be placed **at the end** of your response, **top-level**, and not nested within other tags.
- Always adhere to this format for the tool use to ensure proper parsing and execution.

String and scalar parameters should be specified as is, while lists and objects should use JSON format. Note that spaces for string values are not stripped. The output is not expected to be valid XML and is parsed with regular expressions.

# General Objective

You accomplish a given task iteratively, breaking it down into clear steps and working through them methodically.

## Task Strategy

1. Analyze the user's request and set clear, achievable sub-goals. Prioritize these sub-goals in a logical order.
2. Start with a concise, numbered, step-by-step plan (e.g., 1., 2., 3.) outlining how you will solve the task before taking any action. Each sub-goal should correspond to a distinct step in your task-solving process.
3. Work through these sub-goals sequentially. After each step, the user may provide tool-use feedback, reflect on the results and revise your plan if needed. If you encounter new information or challenges, adjust your approach accordingly. Revisit previous steps to ensure earlier sub-goals or clues have not been overlooked.
4. You have access to a wide range of powerful tools. Use them strategically to accomplish each sub-goal.

## Tool-Use Guidelines

1. Each step must involve a single tool call, unless the task is already solved. 
2. Before each tool call:
- Briefly summarize and analyze what is currently known.
- Identify what is missing, uncertain, or unreliable.
- Be concise; do not repeat the same analysis across steps.
- Choose the most relevant tool for the current sub-goal, and explain why this tool is necessary at this point.
- Verify whether all required parameters are either explicitly provided or can be clearly and reasonably inferred from context.
- Do not guess or use placeholder values for missing inputs.
- Skip optional parameters unless they are explicitly specified.
3. All tool queries must include full, self-contained context. Tools do not retain memory between calls. Include all relevant information from earlier steps in each query.
4. Avoid broad, vague, or speculative queries. Every tool call should aim to retrieve new, actionable information that clearly advances the task.
5. Even if a tool result does not directly answer the question, extract and summarize any partial information, patterns, constraints, or keywords that can help guide future steps.

## Tool-Use Communication Rules

1. Do not include tool results in your response — the user will provide them.
2. Do not present the final answer until the entire task is complete.
3. Do not mention tool names.
4. Do not engage in unnecessary back-and-forth or end with vague offers of help. Do not end your responses with questions or generic prompts.
5. Do not use tools that do not exist.
6. Unless otherwise requested, respond in the same language as the user's message.
7. If the task does not require tool use, answer the user directly.

# Agent Specific Objective

You are a task-solving agent that uses tools step-by-step to answer the user's question. Your goal is to provide complete, accurate and well-reasoned answers using additional tools.

Before verifying all the statements, **always** use the tool to search for information of the statement.

\end{tcblisting}
\begin{tcblisting}{title={Factual Evaluation},colback=lightgrey,colframe=black,arc=1mm,boxrule=1pt,left=1mm,right=1mm,top=1mm,bottom=1mm,breakable,fontupper=\tiny\ttfamily,listing only,listing engine=listings,listing options={breaklines,breakautoindent=false,breakindent=0pt,keepspaces,tabsize=4,literate={“}{"}{1} {”}{"}{1} {‘}{'}{1} {’}{'}{1} {—}{--}{1} {–}{-}{1} { }{ }{1}
    }}
Your task is to perform factual verification of a [part] from an LLM-generated report.
The report is produced in response to a [user query], and both the [user query] and the [part] are provided above.

Task requirements:  
1. Use retrieval tools to fact-check key statements in the [part], collecting evidence from reliable sources whenever possible.
2. You should focus on validating key statements, especially those involving numbers, news, events, dates, locations, or people that can be fact-checked with external sources.  
3. Do not verify every sentence; focus on the core, publicly verifiable claims.
4. For each checked statement, clearly record:  
   - verification: one of [Right, Wrong, Unknown] (choose only one).
     * "Right": the statement matches reliable public information.
     * "Wrong": the statement is contradicted by reliable sources or a statement mixes accurate and inaccurate content.
     * "Unknown": the statement lacks sufficient relevant information, or contains some correct details but other parts can’t be verified.
   - evidence: a list of objects, each containing:  
     * source: the URL of the supporting or contradicting evidence  
     * excerpt: the quoted text or summary from the source that directly relates to the statement  
   - reasoning: a clear explanation of the verification reasoning and process, including how the evidence supports the assigned verification result.

\end{tcblisting}
\begin{tcblisting}{title={Factual Evaluation Summary},colback=lightgrey,colframe=black,arc=1mm,boxrule=1pt,left=1mm,right=1mm,top=1mm,bottom=1mm,breakable,fontupper=\tiny\ttfamily,listing only,listing engine=listings,listing options={breaklines,breakautoindent=false,breakindent=0pt,keepspaces,tabsize=4,literate={“}{"}{1} {”}{"}{1} {‘}{'}{1} {’}{'}{1} {—}{--}{1} {–}{-}{1} { }{ }{1}
    }}
This is a direct instruction to you (the assistant), not the result of a tool call.
    
We are now ending this session, and your conversation history will be deleted. 
You must NOT initiate any further tool use. This is your final opportunity to report 
*all* of the information gathered during the session.

Summarize the above conversation, and output strictly in the following JSON format, with no additional text, explanations, or casual remarks:

{
  "core_state": [
    {
      "statement": "The statement to be verified",
      "verification": "Right / Wrong / Unknown",
      "evidence": [
        {
          "source": "Evidence source URL or reference",
          "excerpt": "Quoted text or summary from the evidence"
        }
      ],
      "reasoning": "Explanation of the verification reasoning and process"
    },
    ...
  ]
}

Requirements:
1. Do not add any output other than the JSON object.
2. Each `evidence` entry must always include both `source` and `excerpt`; do not omit either field.
3. The `source` and `excerpt` must always be included if content is provided.
4. If multiple pieces of evidence exist, include all of them in the `evidence` array.

\end{tcblisting}

\section{Task Construction Details}
\label{apd:DR task}

Our automated deep research task construction pipeline begins with a predefined set of empirical domains. For each domain, we generate multiple personas with clearly defined roles and well-specified backgrounds. Given each persona, we then construct corresponding deep research tasks tailored to that persona. These tasks are subsequently filtered using an LLM-based judge according to predefined criteria.

Next, we generate a no-search baseline for each task and evaluate its performance, discarding tasks that can already be answered well by the baseline alone. Throughout the pipeline, GPT-5-mini is used as the base LLM with default API settings. In total, we define 10 fixed domains, generate 5 personas per domain, and create 4 related deep research tasks for each persona.

\begin{tcblisting}{title={Persona Generation Prompt},colback=lightgrey,colframe=black,arc=1mm,boxrule=1pt,left=1mm,right=1mm,top=1mm,bottom=1mm,breakable,fontupper=\tiny\ttfamily,listing only,listing engine=listings,listing options={breaklines,breakautoindent=false,breakindent=0pt,keepspaces,tabsize=4,literate={“}{"}{1} {”}{"}{1} {‘}{'}{1} {’}{'}{1} {—}{--}{1} {–}{-}{1} { }{ }{1}
    }}
You are a persona generator. For the domain "{{domain}}", generate 5 distinct personas.

Requirements:
1. Personas can be ordinary individuals (e.g., students, hobbyists, employees, community members) or domain experts.
2. Each persona must include a clear background: education, career path, interests, or life experience (100-150 words).
3. Each persona should have specific interests or responsibilities related to the domain.

Example subdomains:
Domain: Software Development
Subdomain: Artificial Intelligence/Machine Learning Development & Tools

Domain: Science & Technology
Subdomain: Large Language Models 
Subdomain: Artificial Intelligence & Machine Learning

Domain: Sports & Fitness
Subdomain: Professional Sports (Football, Basketball, etc.)

Strictly output the following JSON only, with no extra text:
{
  "domain": "{{domain}}",
  "personas": [
    {
      "name": "First persona name",
      "role": "Role or identity (e.g., university student, software engineer, policy advisor)",
      "affiliation": "Affiliation/Organization/Community (if applicable)",
      "background": "Detailed background, including education, work/interest history, and relevant experience (100-150 words)",
      "subdomain": "subdomain within the domain",
    },
    {
      "name": "Second persona name",
      "role": "Role or identity",
      "affiliation": "Affiliation/Organization/Community (if applicable)",
      "background": "Detailed background (100-150 words)",
      "subdomain": "subdomain within the domain",
    },
    ...
  ]

\end{tcblisting}

\begin{tcblisting}{title={Deep Research Generation Prompt},colback=lightgrey,colframe=black,arc=1mm,boxrule=1pt,left=1mm,right=1mm,top=1mm,bottom=1mm,breakable,fontupper=\tiny\ttfamily,listing only,listing engine=listings,listing options={breaklines,breakautoindent=false,breakindent=0pt,keepspaces,tabsize=4,literate={“}{"}{1} {”}{"}{1} {‘}{'}{1} {’}{'}{1} {—}{--}{1} {–}{-}{1} { }{ }{1}
    }}
You are a deep research query designer. Based on the persona profile below, generate 4 realistic deep research queries that the persona would like to ask, which MUST rely on deep web search, reasoning, and information synthesis to complete.

Persona information:
- Name: {{persona_name}}
- Role: {{persona_role}}
- Affiliation: {{persona_affiliation}}
- Background: {{persona_background}}
- Subdomain: {{subdomain}}

Query requirements:
1. Each query must require multiple rounds of web search (at least 2 rounds from different perspectives).
2. Each query must integrate information from multiple credible sources (e.g., academic papers, industry reports, news articles, policy documents, statistics, online forums). ]
3. Cover latest developments, data analysis, trend assessment, comparisons, or case studies — aligned with the persona’s role and domain interest.
4. Complexity should be proportional to the persona’s profile:
   - For experts: advanced analytical or strategic-level queries.
   - For non-experts: practical, decision-making, or problem-solving queries with substantial information gathering.
5. 70% of queries should include explicit time constraints (e.g., "from January 2024 to August 2025", "before 2025", "as of 2025-08-05").
6. Each query must be concrete, specific, and have clear deliverables.
7. Queries must reflect the persona’s realistic needs and challenges.
8. Length: 10-50 words per query.
9. It would be better if query is **tied to a current or emerging hot topic** (e.g., stablecoin regulations in May 2025, AI safety debates, semiconductor export bans) that requires synthesis of conflicting perspectives and time-sensitive evidence.
10. Encourage queries that involve **controversial viewpoints, cross-country comparisons, or forward-looking predictions** (e.g., market impacts, regulatory outcomes, adoption scenarios).

Example queries:
1. My son is about to start his university applications in 2025 for postgraduates but he's still uncertain about both his major and which universities to apply to. Could you help me find the top five universities in each of the five broad subjects from the QS World University Rankings by Subject 2025, and also check their standings in the QS World University Rankings 2025 and the Times Higher Education World University Rankings 2025? And I need the home page of the university's official website, standard application deadline for regular decision as well as the application fee without the fee waiver.
2. Please provide a detailed explanation of the differences and connections between Google's recently released A2A protocol and the MCP protocol. Furthermore, elaborate on the innovative aspects of the A2A protocol and the specific problems it is designed to address.
3. What are the investment philosophies of Duan Yongping, Warren Buffett, and Charlie Munger?
4. How did discussions on stablecoin regulations evolve across US, EU, and Asia in May 2025, and what are the projected implications for financial stability and crypto adoption by 2026?

Strictly output the following JSON only, with no extra text:
{
  "persona_name": "persona_name",
  "tasks": [
    {
      "task_id": "task_1",
      "deep_research_query": "Deep research query (10-50 words)",
      "key_challenges": "Main challenges and why deep search is required",
      "expected_search_rounds": 4,
      "time_sensitivity": true,
      "time_constraint": "Explicit time constraint description"
    },
    {
      "task_id": "task_2",
      "deep_research_query": "Deep research query (10-50 words)",
      "key_challenges": "Main challenges",
      "expected_search_rounds": 3,
      "time_sensitivity": false,
      "time_constraint": null
    },
    ...
  ]
}

\end{tcblisting}

\begin{tcblisting}{title={Task Qualification Filter Prompt},colback=lightgrey,colframe=black,arc=1mm,boxrule=1pt,left=1mm,right=1mm,top=1mm,bottom=1mm,breakable,fontupper=\tiny\ttfamily,listing only,listing engine=listings,listing options={breaklines,breakautoindent=false,breakindent=0pt,keepspaces,tabsize=4,literate={“}{"}{1} {”}{"}{1} {‘}{'}{1} {’}{'}{1} {—}{--}{1} {–}{-}{1} { }{ }{1}
    }}
You are a deep research query analysis expert. Evaluate whether the following task truly requires deep web search and information synthesis to complete.

Query: {{deep_research_query}}
Persona background: {{persona_background}}
Query challenges: {{key_challenges}}

Evaluation criteria:
1. Requires up-to-date information that cannot be accurately answered using pre-2023 knowledge alone.
2. Requires cross-verification and integration of multiple credible sources (e.g., academic papers, news, industry reports, policy documents, statistics).
3. Requires multi-angle, multi-layered deep investigation.
4. Complexity matches the persona’s role and capabilities:
   - For experts: advanced analytical or strategic-level complexity.
   - For non-experts: realistic but still substantial research complexity beyond casual searching.


Strictly output the following JSON only, with no extra text:
{
  "needs_deep_research": (true/false),
  "confidence_score": (0~1),
  "reasoning": "Detailed rationale explaining why deep search is needed and what information collection and synthesis are required (100-150 words)",
  "search_complexity": ("High"/"Medium"/"Low"),
  "information_sources_needed": ["academic papers", "news", "technical reports", "market data", "policy documents"],
  "latest_info_required": (true/false),
  "cross_domain_integration": (true/false)
}

\end{tcblisting}

\begin{tcblisting}{title={Search Necessity Baseline Prompt},colback=lightgrey,colframe=black,arc=1mm,boxrule=1pt,left=1mm,right=1mm,top=1mm,bottom=1mm,breakable,fontupper=\tiny\ttfamily,listing only,listing engine=listings,listing options={breaklines,breakautoindent=false,breakindent=0pt,keepspaces,tabsize=4,literate={“}{"}{1} {”}{"}{1} {‘}{'}{1} {’}{'}{1} {—}{--}{1} {–}{-}{1} { }{ }{1}
    }}
Based solely on your existing knowledge, without using any external search tools, answer the following query. Provide the best answer you can.

Persona background: {{persona_background}}

Query: {{deep_research_query}}

Requirements:
1. Provide as detailed and complete an answer as you can, considering the persona's role, knowledge scope, and domain interest.
2. If some information is uncertain or may be outdated, clearly state this.
3. Do not fabricate specific data, dates, or facts you are not confident about.
4. Ensure logical structure and clear organization.
5. Demonstrate appropriate depth of understanding for the persona's role:
   - For experts: in-depth analysis with technical or strategic insights.
   - For non-experts: practical, clear, and well-explained reasoning.
6. If the query involves recent developments, acknowledge any knowledge cutoff limitations.

Please provide your best possible answer:

\end{tcblisting}

\begin{tcblisting}{title={Search Necessity Assessment Prompt},colback=lightgrey,colframe=black,arc=1mm,boxrule=1pt,left=1mm,right=1mm,top=1mm,bottom=1mm,breakable,fontupper=\tiny\ttfamily,listing only,listing engine=listings,listing options={breaklines,breakautoindent=false,breakindent=0pt,keepspaces,tabsize=4,literate={“}{"}{1} {”}{"}{1} {‘}{'}{1} {’}{'}{1} {—}{--}{1} {–}{-}{1} { }{ }{1}
    }}
You are an answer quality evaluator for deep research queries. Assess the quality and completeness of the following answer for the given query.

Original query: {{deep_research_query}}
Query challenges: {{key_challenges}}
Persona background: {{persona_background}}
No-search answer: {{baseline_answer}}

Evaluation dimensions:
1. Accuracy: Are statements reliable and free of obvious errors?
2. Completeness: Does it cover the core requirements and key aspects of the query?
3. Depth: Is the analysis sufficiently deep and appropriate for the persona’s role?
4. Timeliness: Could information be outdated? Does it acknowledge time limitations?
5. Professionalism: Does it meet the expected standards for the persona’s role and demonstrate relevant insight?
6. Structure: Is the answer well-organized and logical?

Special focus:
- If the task involves time requirements such as "latest", "recent", or "current", assess whether the answer meets them.
- If the task requires multi-source synthesis, assess whether a single knowledge source is sufficient.
- Assess whether additional search is truly necessary to provide a better answer, based on the persona’s role and context.

Strictly output the following JSON only, with no extra text:
{
  "overall_quality": "low",
  "quality_score": 0.3,
  "completeness_score": 0.4,
  "accuracy_score": 0.7,
  "depth_score": 0.2,
  "timeliness_score": 0.1,
  "accuracy_concerns": "Specific accuracy concerns or issues",
  "missing_aspects": "Important aspects and information that are missing",
  "outdated_info": "Potentially outdated content",
  "requires_search": true,
  "search_necessity_reasoning": "Detailed reasoning for why search is needed and what key information is missing (150-200 words)"
}

\end{tcblisting}

\section{Human Study}
\UseRawInputEncoding
\subsection{Correct Examples}
\label{apd:example agent right}

\tcbset{
  factbox/.style={
    enhanced,
    breakable,
    colback=white,
    colframe=black!70,
    boxrule=0.8pt,
    arc=2pt,
    left=6pt,
    right=6pt,
    top=6pt,
    bottom=6pt,
    fonttitle=\bfseries,
  }
}

\begin{tcolorbox}[
  factbox,
  title={Case 1}
]

\textbf{Statement}

DeepSeek V3 was trained for approximately \$5.6 million using 2,788M H800 GPU hours.

\vspace{0.6em}
\textbf{Model Response}

\textbf{Verification:} Wrong.

\textbf{Reasoning:}  
The arXiv paper reports 2.788M H800 GPU hours (i.e., approximately 2.788 million hours). The statement incorrectly uses ``2,788M,'' which implies 2,788 million (i.e., 2.788 billion) GPU hours, introducing a three-orders-of-magnitude unit error. In addition, the \$5.6M figure is a community-derived estimate based on assumed GPU hourly costs, rather than a training cost explicitly reported in the paper. Therefore, the claim as written is incorrect.

\textbf{Evidence:}
\begin{itemize}
  \item \emph{arXiv}: ``Despite its excellent performance, DeepSeek-V3 requires only 2.788M H800 GPU hours for its full training.''
  \item \emph{Medium}: ``2.788M H800 GPU hours (≈\$5.6M at \$2/h) ...''
\end{itemize}

\vspace{0.6em}
\textbf{Human Annotation}

Right

\vspace{0.6em}
\textbf{Final Judge}

The model judgment is \textbf{correct}. Although the discrepancy appears minor, the original statement mistakenly interprets ``2.788M'' as ``2,788M,'' resulting in a thousand-fold overestimation of training compute. This unit-level error materially changes the factual meaning of the claim.

\end{tcolorbox}

\begin{tcolorbox}[
  factbox,
  title={Case 2}
]

\textbf{Statement}

The FIBA Europe Coaching Certificate (FECC), a two-year, mentorship-intensive program run by elite European coaches, is specifically designed to elevate the knowledge base of coaches working with players aged 15--20.

\vspace{0.6em}
\textbf{Model Response}

\textbf{Verification:} Unknown.

\textbf{Reasoning:}  
The FIBA Europe Coaching Certificate (FECC) is a real and highly regarded coaching program focused on the development of youth coaches in Europe. Publicly available sources confirm its prestige and its emphasis on developing coaches working with young players. However, descriptions of the program vary regarding its duration, with some sources referring to a three-year structure, and publicly available materials do not consistently substantiate the precise phrasing ``two-year, mentorship-intensive'' nor explicitly specify a target player age range of 15--20. Due to these inconsistencies and insufficient public detail, the exact claim cannot be fully verified.

\textbf{Evidence:}
\begin{itemize}
  \item \emph{FIBA Europe}: ``The FIBA Europe Coaching Certificate (FECC) program has established its position as one of the most prestigious coaching programs in European sport.''
  \item \emph{FIBA Europe Coaching}: ``The graduates from the FECC will receive an extremely rare and valuable qualification and they are FIBA Europe approved coaches for developing young players.''
  \item \emph{African Sports Monthly}: ``The exclusive three-year coaching program, which focuses on the development of young players, has a limited intake of 60 up-and-coming European coaches...''
\end{itemize}

\vspace{0.6em}
\textbf{Human Annotation}

Right

\vspace{0.6em}
\textbf{Final Judge}

The model judgment is \textbf{correct}. A review of publicly available information on the FIBA Europe Coaching Certificate (FECC) confirms that most aspects of the statement are broadly accurate. However, official descriptions do not explicitly specify a player age range of 15--20, and the program duration is described inconsistently across sources. As a result, the claim cannot be fully verified and is appropriately classified as Unknown.

\end{tcolorbox}

\begin{tcolorbox}[
  factbox,
  title={Correct Case 3}
]

\textbf{Statement}

European Union: Dutch and EU-wide measures broadened controls on advanced lithography and semiconductor manufacturing equipment (e.g., immersion DUV tools, EUV pellicles).

\vspace{0.6em}
\textbf{Model Response}

\textbf{Verification:} Unknown.

\textbf{Reasoning:}  
Public documentation from the Dutch government and ASML confirms that the Netherlands expanded export license requirements on advanced semiconductor manufacturing equipment, notably including immersion DUV lithography tools, in September 2024. However, no publicly available primary EU regulatory documents were found that explicitly demonstrate an EU-wide measure adding EUV pellicles to the EU dual-use export control list by the relevant time frame. Because the statement jointly attributes such controls to both Dutch national measures and EU-wide actions, including EUV pellicles, the combined claim cannot be fully verified.

\textbf{Evidence:}
\begin{itemize}
  \item \emph{ASML Press Release}: ``The Dutch government today published an updated license requirement regarding the export of immersion DUV semiconductor equipment.''
  \item \emph{Government of the Netherlands}: ``As of 7 September 2024, the national export control measure applicable to advanced semiconductor manufacturing equipment will be expanded.''
\end{itemize}

\vspace{0.6em}
\textbf{Human Annotation}

Right

\vspace{0.6em}
\textbf{Final Judge}

The model judgment is \textbf{correct}. A review of relevant public sources confirms that export controls were expanded at the Dutch national level, while no clear evidence supports an EU-wide regulatory action specifically extending controls to EUV pellicles. The original statement implies coordinated Dutch and EU-wide expansion of such controls, which is not fully supported by available evidence; therefore, the claim is appropriately classified as Unknown.

\end{tcolorbox}

\subsection{Incorrect Examples}
\label{apd:example agent wrong}

We present two examples in which overly strict verification leads to incorrect judgments; however, it should be noted that such cases occur only occasionally.

\begin{tcolorbox}[
  factbox,
  title={Incorrect Case 1}
]

\textbf{Statement}

AI-based virtual triage is effective at identifying individuals with potentially life-threatening conditions who initially did not intend to seek professional care.

\vspace{0.6em}
\textbf{Model Response}

\textbf{Verification:} Unknown.

\textbf{Reasoning:}  
There are individual examples of digital triage flagging urgent conditions, but the overall performance of symptom checkers and AI-based triage tools varies across systems and clinical settings. The broad and unconditional claim that AI virtual triage is effective in this role lacks a single, universal supporting source. Therefore, the claim is classified as Unknown.

\textbf{Evidence:}
\begin{itemize}
  \item \emph{BMJ}: ``Evaluations of symptom checkers show widely variable performance; some identify urgent conditions, but overall sensitivity and safety vary by tool and condition.''
  \item \emph{PMC}: ``AI-based triage systems show promise for improving emergency department efficiency but require rigorous multi-center validation and standardized outcome reporting.''
\end{itemize}

\vspace{0.6em}
\textbf{Human Annotation}

Right

\vspace{0.6em}
\textbf{Final Judge}

The model judgment is \textbf{incorrect}.  
Although the model identified concrete examples supporting the claim, it applied an overly strict verification standard by requiring a universal supporting source. The evidence it retrieved is already sufficient to substantiate the statement, and therefore the claim should be considered correct.

\end{tcolorbox}

\begin{tcolorbox}[
  factbox,
  title={Incorrect Case 2}
]

\textbf{Statement}

Technology procurement for basic AI triage systems ranges from \$50{,}000 to \$200{,}000.

\vspace{0.6em}
\textbf{Model Response}

\textbf{Verification:} Unknown.

\textbf{Reasoning:}  
Industry and vendor analyses report wide and overlapping ranges for AI implementation costs in healthcare, spanning from tens of thousands to well over one million dollars depending on scale and complexity. Although the 50k to 200k range is plausible for certain deployments, it is not established as a general or authoritative standard in peer-reviewed or governmental sources. Therefore, the specific figure cannot be conclusively verified and is classified as Unknown.

\textbf{Evidence:}
\begin{itemize}
  \item \emph{Biz4Group}: ``AI in healthcare applications ranges from \$50K–\$1M+ depending on scale and complexity.''
  \item \emph{ITRex Group}: ``The costs of implementing AI in healthcare range from \$40{,}000 for simple AI functionality to \$100{,}000 and much more for a comprehensive solution.''
  \item \emph{Aalpha}: ``The cost of implementing AI in healthcare can range from \$50{,}000 to \$300{,}000 for small clinics.''
  \item \emph{Provided Part}: ``Procurement … can range from \$50{,}000 to \$200{,}000 for basic AI triage systems depending on customization needs.''
\end{itemize}

\vspace{0.6em}
\textbf{Human Annotation}

Right

\vspace{0.6em}
\textbf{Final Judge}

The model judgment is \textbf{incorrect}.  
The original source explicitly reports this numerical range, but the model dismissed the claim by treating the specific figure as inappropriate and applying an overly strict verification criterion. In fact, the statement is supported by the source and should be judged as correct.

\end{tcolorbox}

\section{Examples}
\label{apd:sec example}

\UseRawInputEncoding

\begin{tcblisting}{title={Task-Specific Dimension Example 1},colback=lightgrey,colframe=black,arc=1mm,boxrule=1pt,left=1mm,right=1mm,top=1mm,bottom=1mm,breakable,fontupper=\tiny\ttfamily,listing only,listing engine=listings,listing options={breaklines,breakautoindent=false,breakindent=0pt,keepspaces,tabsize=4,literate={“}{"}{1} {”}{"}{1} {‘}{'}{1} {’}{'}{1} {—}{--}{1} {–}{-}{1} { }{ }{1} {…}{{...}}1
    }}

{"topic": "Transportation", "language": "en", "prompt": "Compare e‑scooter regulatory frameworks, safety outcomes, and injury trends in US, EU, and China from January 2023 to August 2025; deliver harmonized policy recommendations, five measurable safety metrics, and data sources."}

  "dimensions": [
    {
      "meta_dimension_name": "Policy Pragmatism",
      "definition": "Assesses the real-world viability and implementability of the harmonized policy recommendations, evaluating their adaptability to the distinct political, economic, and cultural contexts of the US, EU, and China."
    },
    {
      "meta_dimension_name": "Comparative Synthesis",
      "definition": "Evaluates the quality of the cross-jurisdictional analysis, measuring how effectively the report integrates disparate regulatory frameworks and safety data into a coherent, unified narrative, rather than merely juxtaposing separate regional summaries."
    },
    {
      "meta_dimension_name": "Metric Utility",
      "definition": "Measures the quality of the proposed safety metrics, evaluating if they are genuinely comparable, measurable, and actionable across the diverse data collection ecosystems of the US, EU, and China."
    }
  ]

\end{tcblisting}

\begin{tcblisting}{title={Task-Specific Dimension Example 2},colback=lightgrey,colframe=black,arc=1mm,boxrule=1pt,left=1mm,right=1mm,top=1mm,bottom=1mm,breakable,fontupper=\tiny\ttfamily,listing only,listing engine=listings,listing options={breaklines,breakautoindent=false,breakindent=0pt,keepspaces,tabsize=4,literate={“}{"}{1} {”}{"}{1} {‘}{'}{1} {’}{'}{1} {—}{--}{1} {–}{-}{1} { }{ }{1} {…}{{...}}1
    }}
    
{"topic": "Health", "language": "en", "prompt": "Assess nutritional quality, additive profiles, sodium levels, and 'ultra-processed' classification of top plant-based meat products across US and EU as of 2025-08-15; deliver comparative tables and public-health recommendations."}

  "dimensions": [
    {
      "meta_dimension_name": "Classification Rigor",
      "definition": "Assesses the clarity, consistency, and scientific validity of applying the 'ultra-processed' classification framework (e.g., NOVA) to the products. This is crucial because the classification is a central, non-trivial analytical task of the query, not just a simple data point."
    },
    {
      "meta_dimension_name": "Cross-Regional Synthesis",
      "definition": "Evaluates the depth of the comparative analysis between US and EU markets, assessing not just the presentation of data from both regions, but the synthesis of differences in product formulation, nutritional standards, and regulatory landscapes that explain the observed data."
    },
    {
      "meta_dimension_name": "Recommendation Actionability",
      "definition": "Measures how specific, practical, and evidence-based the public-health recommendations are, and how directly they are derived from the report's comparative findings on nutrition, additives, and processing, rather than being generic health advice."
    }
  ]




\end{tcblisting}

\begin{tcblisting}{title={Task-Specific Dimension Example 3},colback=lightgrey,colframe=black,arc=1mm,boxrule=1pt,left=1mm,right=1mm,top=1mm,bottom=1mm,breakable,fontupper=\tiny\ttfamily,listing only,listing engine=listings,listing options={breaklines,breakautoindent=false,breakindent=0pt,keepspaces,tabsize=4,literate={“}{"}{1} {”}{"}{1} {‘}{'}{1} {’}{'}{1} {—}{--}{1} {–}{-}{1} { }{ }{1} {…}{{...}}1
    }}

{"topic": "Sports & Fitness", "language": "en", "prompt": "Compare community youth basketball coaching curricula (US AAU/USAB, Spain club youth pathway, China school/club models): skill retention, coach training, and transition rates; deliver a scalable municipal implementation toolkit."}

  "dimensions": [
    {
      "meta_dimension_name": "Implementation Viability",
      "definition": "Assesses the practicality, scalability, and adaptability of the proposed municipal toolkit. This evaluates whether the deliverable includes actionable templates, realistic budget considerations, and clear guidance for tailoring the program to diverse local contexts (e.g., city size, funding levels, existing infrastructure)."
    },
    {
      "meta_dimension_name": "Comparative Synthesis",
      "definition": "Evaluates the ability to analyze each country's basketball development system within its unique socio-cultural and structural context, and to synthesize these disparate findings into transferable principles. This goes beyond a side-by-side description to assess the quality of drawing actionable lessons from one system for application in another."
    }
  ]




\end{tcblisting}

\begin{tcblisting}{title={Factual Evaluation Example 1},colback=lightgrey,colframe=black,arc=1mm,boxrule=1pt,left=1mm,right=1mm,top=1mm,bottom=1mm,breakable,fontupper=\tiny\ttfamily,listing only,listing engine=listings,listing options={breaklines,breakautoindent=false,breakindent=0pt,keepspaces,tabsize=4,literate={“}{"}{1} {”}{"}{1} {‘}{'}{1} {’}{'}{1} {—}{--}{1} {–}{-}{1} { }{ }{1} {…}{{...}}1
    }}

{
  "statement": "China produces two-thirds of the world’s consumer electronics products.",
  "verification": "Wrong",
  "evidence": [
    {
      "source": "https://www.ey.com/en_cn/consumer-products/ey-market-trends-report-2024-en.pdf",
      "excerpt": "China’s dominant position in consumer electronics exports is undeniable, with its global share reaching an impressive 42% in 2022."
    },
    {
      "source": "https://rhg.com/research/china-and-the-future-of-global-supply-chains/",
      "excerpt": "We review China’s role in consumer electronics… China is the world’s largest electronics manufacturing hub."
    }
  ],
  "reasoning": "Authoritative industry analyses indicate that China is indeed the world’s largest manufacturing hub for consumer electronics, but recent estimates typically place its share at around 40–45%, rather than approximately 66%. The available sources do not support the claim that China produces ‘two-thirds’ of global consumer electronics."
}


\end{tcblisting}

\begin{tcblisting}{title={Factual Evaluation Example 2},colback=lightgrey,colframe=black,arc=1mm,boxrule=1pt,left=1mm,right=1mm,top=1mm,bottom=1mm,breakable,fontupper=\tiny\ttfamily,listing only,listing engine=listings,listing options={breaklines,breakautoindent=false,breakindent=0pt,keepspaces,tabsize=4,literate={“}{"}{1} {”}{"}{1} {‘}{'}{1} {’}{'}{1} {—}{--}{1} {–}{-}{1} { }{ }{1} {…}{{...}}1
    }}

{
  "statement": "China / SMIC developed 7nm chips without EUV equipment and has further advanced to mass production at 5nm.",
  "verification": "Wrong",
  "evidence": [
    {
      "source": "https://www.bloomberg.com/news/features/2023-09-04/look-inside-huawei-mate-60-pro-phone-powered-by-made-in-china-chip",
      "excerpt": "Huawei’s Mate 60 Pro is powered by the new Kirin 9000S chip … manufactured in China by SMIC … the processor is the first product made using SMIC’s most advanced 7-nanometer process."
    },
    {
      "source": "https://americanaffairsjournal.org/2024/02/a-new-era-for-the-chinese-semiconductor-industry-beijing-responds-to-export-controls/",
      "excerpt": "Assuming Beijing’s new semiconductor industrial policies achieve significant results … by 2025 SMIC could significantly expand capacity at the 5/6/7 nm nodes … in the most optimistic scenario … SMIC may obtain a prototype EUV system for testing in 2025."
    }
  ],
  "reasoning": "The evidence supports the factual claim that SMIC has produced 7nm-class chips using advanced DUV techniques without EUV lithography (e.g., the Kirin 9000S used in the Mate 60). However, during the period from 2023 to mid-2024, no reliable sources confirm that SMIC has achieved stable mass production at the 5nm node; discussions of 5nm progress remain speculative or forward-looking. Because the statement combines a verified fact (7nm without EUV) with an unverified forward-looking claim (5nm mass production), and under the verification rule that mixed accuracy results in an overall negative judgment, the statement is rated as Wrong."
}



\end{tcblisting}

\begin{tcblisting}{title={Factual Evaluation Example 3},colback=lightgrey,colframe=black,arc=1mm,boxrule=1pt,left=1mm,right=1mm,top=1mm,bottom=1mm,breakable,fontupper=\tiny\ttfamily,listing only,listing engine=listings,listing options={breaklines,breakautoindent=false,breakindent=0pt,keepspaces,tabsize=4,literate={“}{"}{1} {”}{"}{1} {‘}{'}{1} {’}{'}{1} {—}{--}{1} {–}{-}{1} { }{ }{1} {…}{{...}}1
    }}

{
  "statement": "Surveys show that the public holds a low level of optimism about the social impact of artificial intelligence, with only 27% of Americans believing that AI has a positive effect on science and medicine.",
  "verification": "Wrong",
  "evidence": [
    {
      "source": "https://www.pewresearch.org/internet/2025/04/03/how-the-us-public-and-ai-experts-view-artificial-intelligence/",
      "excerpt": "There are also large differences in views of AI’s impact on the economy, health care, education and the arts. … Health care is the area where the public is most optimistic about AI’s impact. Experts are much more likely than the public to say it will have a positive effect, with a gap of as much as 40 percentage points (84% vs. 44%)."
    },
    {
      "source": "https://www.pewresearch.org/wp-content/uploads/sites/20/2025/04/pi_2025.04.03_us-public-and-ai-experts_report.pdf",
      "excerpt": "These surveys reveal deep divides and some common ground on AI issues. Experts are far more positive than the public about AI’s potential … The public is more likely to say AI will harm them (43%) than benefit them (24%). … Experts are more likely than the public to say AI will have a very or somewhat positive effect on the United States over the next 20 years (56% vs. 17%)."
    }
  ],
  "reasoning": "The Pew Research Center reports do not show that '27% of Americans believe AI has a positive effect on science and medicine.' Instead, the reports indicate that 44% of U.S. adults believe AI will have a positive impact on health care, and 17% believe it will have a positive impact on the United States overall. Therefore, the '27%' figure cited in the statement is not supported by the Pew findings and is incorrect."
}




\end{tcblisting}


\end{document}